\documentclass[journal]{IEEEtran}

\usepackage[style=nejm, 
citestyle=numeric-comp,
sorting=none]{biblatex}
\usepackage{graphicx}

\graphicspath{ {./figures/} }
\usepackage{subcaption}
\usepackage{amsmath}
\usepackage{lineno}
\usepackage[russian,english]{babel}
\usepackage[utf8]{inputenc}
\usepackage{amssymb}
\usepackage{booktabs}
\usepackage{multirow}
\usepackage{hyperref}

\ifCLASSINFOpdf
\else
\fi

\begin{document}

\title{{\fontsize{16pt}{16pt}\selectfont 
\textbf{Learning Multimodal AI Algorithms for Amplifying Limited User Input \\ 
into High-dimensional Control Space}
}}

\author{Ali~Rabiee,
        Sima~Ghafoori,
        MH~Farhadi,
        Robert~Beyer,
        Xiangyu~Bai,
        David J Lin,
        Sarah~Ostadabbas,
        and~Reza~Abiri
        
\thanks{A. Rabiee, S. Ghafoori, MH Farhadi, Robert Beyer, and R. Abiri are with the Department
of Electrical, Computer, and Biomedical Engineering, University of Rhode Island, Kingston,
RI, USA. Xiangyu Bai and Sarah Ostadabbas are with the Department of Electrical and Computer Engineering, Northeastern University, Boston, MA, USA. David J Lin is with the Department of Neurology, Harvard Medical School, Boston, MA, USA.
David J Lin, Sarah Ostadabbas, and Reza Abiri (corresponding author) are senior authors.}
}

\maketitle

\begin{abstract}
Current invasive assistive technologies (e.g., brain-controlled prostheses) are designed to infer high-dimensional motor control signals from severely paralyzed patients. However, they face significant challenges, including public acceptance, limited longevity, and barriers to commercialization. Meanwhile, noninvasive alternatives often rely on artifact-prone signals, require lengthy user training, and struggle to deliver robust high-dimensional control for dexterous tasks. To address these issues, this study introduces a novel human-centered multimodal AI approach as intelligent compensatory mechanisms for lost motor functions that could potentially enable patients with severe paralysis to control high-dimensional assistive devices, such as dexterous robotic arms, using limited and noninvasive inputs ( e.g., head motions). In contrast to the current state-of-the-art (SoTA) noninvasive approaches, our context-aware, multimodal shared-autonomy framework integrates deep reinforcement learning algorithms to blend limited low-dimensional user input with real-time environmental perception, enabling adaptive, dynamic, and intelligent interpretation of human intent for complex dexterous manipulation tasks, such as pick-and-place. The results from our ARAS (Adaptive Reinforcement learning for Amplification of limited inputs in Shared autonomy) trained with synthetic users over 50,000 computer simulation episodes demonstrated the first successful implementation of the proposed closed-loop human-in-the-loop paradigm outperforming the SoTA shared autonomy algorithms. Following a zero-shot sim-to-real transfer, ARAS was evaluated on 23 human subjects, demonstrating high accuracy in dynamic intent detection and smooth, stable 3D trajectory control for dexterous pick-and-place tasks. ARAS user study achieved a high task success rate of 92.88\%, with short completion times comparable to those of SoTA invasive assistive technologies (e.g., brain-controlled robotic arms). The ARAS codes can be accessed at: 
\href{https://github.com/AbiriLab/ARAS}{https://github.com/AbiriLab/ARAS}

\end{abstract}



\IEEEpeerreviewmaketitle

\section{Introduction}

Human-robot collaboration in assistive prostheses and robotics has emerged as a critical field of study, aiming to enhance the independence and quality of life for individuals with motor impairments \cite{de2022artificial, emken2007human, ghafoori2024novel}. This domain intersects robotics, human-computer interaction, and rehabilitation engineering, focusing on developing systems that can effectively augment human capabilities. Key applications within this field include assistive manipulation tasks essential for daily living \cite{somers2024spinal, carver2016impact, abiri2024toward}. For instance, a person with quadriplegia might use a dexterous robotic arm to independently grasp and drink from a cup of water, alleviating the need for constant caregiver assistance \cite{hochberg2012reach}.

Typical robotic manipulators used in assistive contexts require control over at least 7 degrees of freedom (DoF) - three for positioning the end-effector in 3D space, three for its orientation, and one for opening or closing the gripper. This high-dimensional control space demands rich, high-quality input signals from the user \cite{lee2024learning, mohebbi2020human}. To complete such high-dimensional control tasks, previous studies including Hochberg et al. \cite{hochberg2012reach} and our prior work, Natraj et al. \cite{natraj2025sampling} have focused on invasive methods such as intracortical brain-computer interfaces (BCIs), where electrodes are surgically implanted into the brain to directly decode intended dexterous motions. These methods can, in theory, provide high-dimensional control commands by recording high bandwidth neural signals from motor and sensory cortices of the brain. These unimodal approaches have shown promising results in controlling high-dimensional robotic devices in laboratory settings \cite{muelling2017autonomy, downey2016blending,}, however, they imposed significant drawbacks in translational research and commercialization due to the risk of surgical operation, the longevity of signals, the re-calibration requirement, and early fatigue for the users \cite{collinger2013high, silversmith2021plug}. 

\begin{figure}[!ht]
    \centering
    \includegraphics[width=.48\textwidth]{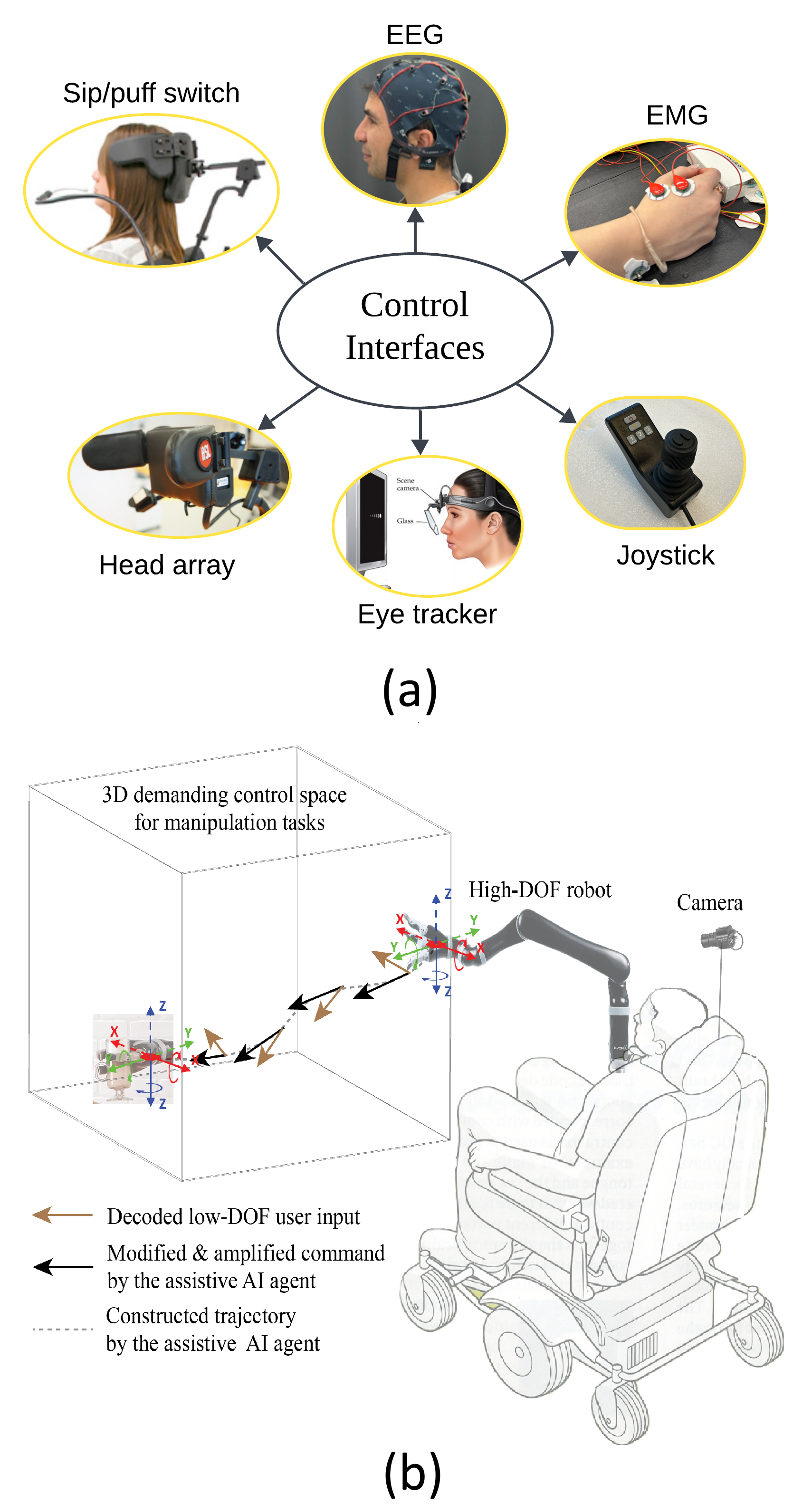}
    \caption{(a) Examples of noninvasive user control interfaces commonly employed in assistive robotics. EEG: Electroencephalography. EMG: Electromyography (b) A conceptual design of an assistive AI-powered robotic manipulation system amplified low-dimensional user inputs for high-DoF tasks.}
    \label{fig:conceptual}
\end{figure}

Given such barriers, non-invasive assistive technologies have gained increasing attention as alternatives to invasive techniques such as implantable BCIs (Figure \ref{fig:conceptual}a). These include widely used Electroencephalography (EEG)-based systems \cite{10782674, ghafoori2024bispectrum, cetera2024classification, xu2019shared, beraldo2022shared}, as well as conventional devices like joysticks, keyboards, and single-switch interfaces that translate residual physical movements—such as finger motions or head tilts—into control signals \cite{cook2002assistive, tai2008review}. However, non-invasive methods still face major hurdles when high-dimensional control is required. Surface signals acquired from the scalp or skin demand artifact-free recordings, which frequently entails lengthy user training to achieve reliable performance \cite{rabiee2024comparative, abiri2019comprehensive, meng2016noninvasive, lafleur2013quadcopter, artemiadis2010emg}. In parallel, motion-based control strategies rely on continuous, dexterous inputs that many individuals with severe motor impairments cannot produce. Instead, these users are often restricted to discrete commands, such as binary left/right signals from a head array or simple on/off inputs from a single-switch device \cite{tai2008review, wolpaw2002brain, mugler2010design}, making it difficult to operate high-dimensional assistive devices.

To address these challenges, we propose Adaptive Reinforcement Learning for Amplification of limited inputs in Shared autonomy (ARAS). ARAS dynamically interprets limited user inputs and amplifies them into high-dimensional control actions (Figure \ref{fig:conceptual}b), leveraging a multimodal deep reinforcement learning architecture that fuses historical user input patterns with real-time environmental perception. This integrated approach ensures adaptive, context-aware assistance by continually updating goal inference in response to changing user intentions and object locations. Through extensive simulation and real-world evaluations, ARAS has demonstrated robust performance, enhancing task efficiency, minimizing cognitive and physical effort, and preserving user control while achieving high success rates in dexterous tasks such as pick-and-place.

Our key contributions include: (1) development of an adaptive reinforcement learning formulation to dynamically infer and amplify user inputs based on historical and real-time data, (2) a multimodal integration approach combining non-invasive user signals with camera-based perception for intent recognition, (3) validation of the proposed framework through extensive simulations and real-world user studies, which show significant improvements in task efficiency, user satisfaction, and success rates compared to baselines, and (4) a publicly available implementation of the ARAS framework to further advance research in shared autonomy for assistive robotics.

\section{Related Works}
A common solution to address the challenge of high-dimensional control is mode-switching approaches \cite{herlant2016assistive, rudigkeit2014towards}. In these systems, users switch between different control modes, each corresponding to a subset of the robot's degrees of freedom or specific functions such as translational mode or rotational mode. For example, a user might switch between modes for controlling the arm's position, orientation, and gripper state. However, mode-switching approaches face several limitations. Frequent mode switches can significantly slow down task execution, especially for complex tasks requiring multiple DoF adjustments \cite{herlant2016assistive}. Users must maintain awareness of the current mode, which is cognitively demanding \cite{carlson2013brain}. The mapping between user inputs and robot actions changes across modes, which can be unintuitive and require substantial training \cite{rudigkeit2014towards}. Therefore, auto mode-switching techniques \cite{ardon2019learning} have been developed to reduce the cognitive load associated with manual mode switching. These systems attempt to automatically change modes based on the context of the task and predicted user intentions. For instance, the system might automatically switch to a fine-control mode when approaching an object to be grasped. While auto mode-switching can alleviate some issues, it introduces new challenges. Automatic mode changes may not always align with user intentions, potentially causing frustration or errors \cite{sellen1992prevention, surale2017experimental}. Predefined modes may not be suitable for all tasks or user preferences, limiting the system's flexibility \cite{ardon2019learning}. Even with auto-switching, users must still anticipate mode changes, which can remain cognitively demanding \cite{carlson2013brain}. 

Context-aware shared autonomy systems aim to reduce the user’s control burden by integrating autonomous behaviors guided by high-level goals or inferred intentions, using environmental information and arbitration functions \cite{hung2013context, sadeghi2024systematic, peer2008human, dragan2013policy}. Prior research in this area typically assumes that the user’s goal is either fully known or can be inferred from a predetermined set of options. These methods often employ Markov Decision Processes (MDPs) when the goal is known \cite{dragan2013policy, javdani2015shared, yow2023shared} and Partially Observable Markov Decision Processes (POMDPs) when the goal is uncertain \cite{nikolaidis2017human, javdani2018shared}. Although some work addresses unknown goals, it usually relies on restrictive assumptions—such as a fixed goal or minimal autonomous assistance \cite{javdani2018shared}—making it difficult to handle dynamic, evolving goals with limited user inputs.

Many context-aware approaches use goal-based strategies. In user-specified goal systems \cite{stepputtis2020language, shafti2019gaze}, individuals convey their targets explicitly, for example via voice \cite{stepputtis2020language}, eye gaze \cite{shafti2019gaze}, or discrete keyboard inputs, sometimes with multiple modalities combined for robustness \cite{muelling2017autonomy}. In goal recognition systems \cite{awais2010human, trick2019multimodal}, the robot infers the user’s intention from contextual cues and initial actions, often using inverse reinforcement learning \cite{ziebart2009planning} or Bayesian inference \cite{best2015bayesian}. More recent methods provide assistance even when no explicit goal is given \cite{javdani2018shared, yow2023shared}, but they still face major hurdles. One issue is the continuum of goals in real-world tasks (for example, determining the precise placement of an object or the exact portion to cut), making it impractical for users to specify exact targets through simple commands \cite{javdani2018shared}, especially if speech is impaired \cite{hazelton2022interventions, silversmith2021plug}. Moreover, explicit goal communication can reduce teamwork effectiveness \cite{goodrich2003seven, green2008human}, and fully autonomous systems often increase user stress \cite{pollak2020stress} without providing a clear advantage in user satisfaction \cite{kim2011autonomy}. Users also tend to respond more negatively to fully autonomous robots than to partially autonomous ones \cite{zlotowski2017can}.

Building on these observations, we note that the complexity of user intent uncertainty, high-dimensional control requirements, the discrete and potentially faulty nature of user inputs, and dynamically evolving tasks remain major hurdles for traditional assistive robotic control strategies \cite{jandaghi2023motion, jandaghi2024composite}. There is a pressing need for solutions that effectively handle these uncertainties while maintaining interpretability \cite{khani2024explainable, afshar2024ibo} and adaptive decision-making to empower users in assistive robotics.
 
\section{Problem Formulation}
\label{sec:formulation}

 We propose a novel model-free reinforcement learning formulation called ARAS.

\begin{figure}[!ht]
\centering
\includegraphics[width=.47\textwidth]{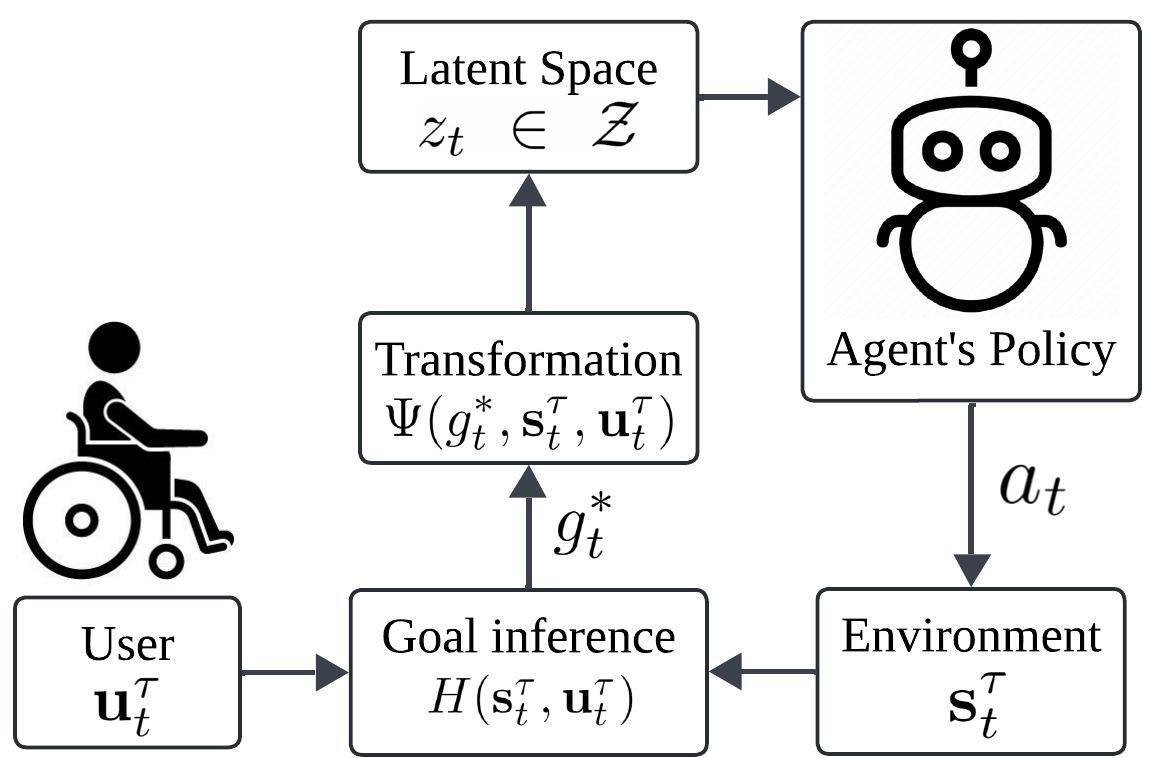}
\caption{ARAS formulation diagram.}
\label{fig:highlevel_diagram}
\end{figure}

Our ARAS formulation extends the traditional POMDP framework to explicitly account for the amplification of limited and low-dimensional user inputs and the dynamic nature of goal inference in shared autonomy scenarios. We consider a model-free reinforcement learning approach by leveraging prior research \cite{reddy2018shared}, thus eliminating the need for a transition function. Such a model can potentially be learned to dynamically infer the user's intent through a more informative latent space representation before determining the robot's actions. As shown in Figure \ref{fig:highlevel_diagram}, we define our formulation as a tuple:

\begin{equation}
(\mathcal{S}, \mathcal{A}, \mathcal{U}, G, H, \mathcal{Z}, R, \gamma),
\label{eq:tuple}
\end{equation} \\
where $\mathcal{S}$ represents the state space of the system, capturing both the robot’s state $x_t^{\text{robot}} \in \mathcal{S}_{\text{robot}}$ and the environmental state $x_t^{\text{env}} \in \mathcal{S}_{\text{env}}$ at time step $t$. The historical information over the previous $\tau$ time steps is incorporated into the system as $\mathbf{s}_t^\tau = \{ s_t, s_{t-1}, \dots, s_{t-\tau+1} \}$, which is essential for tracking dynamic state changes over time.  $\mathcal{A}$ is the action space that defines the set of high-dimensional actions $a_t \in \mathcal{A}$ that the robot can perform. The user input space $\mathcal{U}$ refers to the low-dimensional inputs generated by the user, typically limited in DoF, which guide the robot’s actions. These inputs are collected over time, with the historical user inputs defined as $\mathbf{u}_t^\tau = \{ u_t, u_{t-1}, \dots, u_{t-\tau+1} \}$. \( G \) represents the goal space, which contains the possible goals or intentions the user may have during the task.

The user's goal is dynamically inferred through the function $H(\mathbf{s}_t^\tau, \mathbf{u}_t^\tau)$, which provides a distribution over possible goals in the goal space $G$. To model this inference process more rigorously, we leveraged Bayesian inference \cite{box2011bayesian} for $H$. Given the prior belief over the goals and the current observations (state and user inputs), the goal inference follows Bayes' rule:

\begin{equation}
P(g | \mathbf{s}_t^\tau, \mathbf{u}_t^\tau) = \frac{P(\mathbf{s}_t^\tau, \mathbf{u}_t^\tau | g) P(g)}{\sum_{g' \in G} P(\mathbf{s}_t^\tau, \mathbf{u}_t^\tau | g') P(g')},
\label{eq:bayes}
\end{equation} \\
where $P(g | \mathbf{s}_t^\tau, \mathbf{u}_t^\tau)$ is the posterior distribution over goals given the historical states and user inputs, $P(g)$ is the prior probability of goal $g$, and $P(\mathbf{s}_t^\tau, \mathbf{u}_t^\tau | g)$ is the likelihood function, which represents the probability of observing the current states and inputs given that goal $g$ is the true goal. We define the likelihood function $P(\mathbf{s}_t^\tau, \mathbf{u}_t^\tau | g)$ based on how well the observed state and user inputs align with the expected behavior under goal $g$. Formally, this can be expressed as:

\begin{equation}
P(\mathbf{s}_t^\tau, \mathbf{u}_t^\tau | g) = \prod_{t'=t-\tau+1}^{t} \mathcal{L}(s_{t'}^{\text{robot}}, s_{t'}^{\text{env}}, u_{t'} | g),
\label{eq:likelihood}
\end{equation} \\
where $\mathcal{L}(s_{t'}^{\text{robot}}, s_{t'}^{\text{env}}, u_{t'} | g)$ is a likelihood term that evaluates the alignment between the robot’s state, the environment, the user input, and the goal $g$. This likelihood term can be designed as a Gaussian distribution centered around the expected states and inputs for goal $g$:

\begin{equation}
\mathcal{L}(s_{t'}^{\text{robot}}, s_{t'}^{\text{env}}, u_{t'} | g) = \mathcal{N}((s_{t'}^{\text{robot}}, s_{t'}^{\text{env}}, u_{t'}) | \mu_g, \Sigma_g),
\label{eq:likelihood_gaussian}
\end{equation} \\
where $\mu_g$ represents the expected state and input values for goal $g$, and $\Sigma_g$ is the covariance matrix representing the variability in these observations. Thus, the most probable goal $g_t^* \in G$ at time $t$ is defined as the goal that maximizes the posterior distribution:

\begin{equation}
g_t^* = \text{argmax}_{g \in G} P(g | \mathbf{s}_t^\tau, \mathbf{u}_t^\tau).
\label{eq:max_posterior}
\end{equation} \\

Next, \( \mathcal{Z} \) is the latent space, which is a compressed, task-relevant representation that combines the inferred goal \( g_t^* \), the historical states \( \mathbf{s}_t^\tau \), and the user inputs \( \mathbf{u}_t^\tau \) into a lower-dimensional space. This latent space serves as a more compact form of state representation, designed to facilitate efficient action selection. The transformation into the latent space is achieved through a mapping function \( \Psi \), defined as:

\begin{equation}
z_t = \Psi(g_t^*, \mathbf{s}_t^\tau, \mathbf{u}_t^\tau).
\label{eq:latent_mapping}
\end{equation}

In cases where the confidence in the inferred goals is low, the system may infer a null goal (\( g_t^* = \text{null} \)), indicating uncertainty about the user's intention. Theoretically, in these scenarios, the robot’s actions would primarily follow the user's direct inputs, relying on real-time guidance. The focus would shift toward modifying the potentially noisy and imprecise inputs by considering historical user input data \(\mathbf{u}_t^\tau\). Alternatively, the robot may take a "hold action", pausing to gather more information from the user until there is sufficient confidence to infer a concrete goal. This adaptive behavior ensures the system can handle uncertain or evolving user goals without forcing premature actions.

The agent's policy \( \pi: \mathcal{Z} \rightarrow \mathcal{A} \) is learned off-policy, using a target policy to map the latent representation \( z_t \) to high-DoF robot actions \( a_t \) while maximizing the cumulative reward. The reward function $R(\mathbf{s}_t^\tau, a_t)$ is inspired by human centered and neuro-scientific principles of human reward processing, comprising three main components:

\begin{equation}
R(\mathbf{s}_t^\tau, a_t) = \alpha R_{\text{GP}}(g_t^*, \mathbf{s}_t^\tau, a_t) + \beta R_{\text{IA}}(\mathbf{s}_t^\tau, \mathbf{u}_t^\tau) + \delta R_{\text{TC}}(\mathbf{s}_t^\tau),
\label{eq:reward}
\end{equation} \\
where $R_{\text{GP}}$ (Goal-Progress) reflects intrinsic motivation and progress towards goals, mirroring the progress of task-tracking function of the prefrontal cortex and basal ganglia \cite{haber2010reward}. In our robotic context, this reward promotes movement toward the inferred goal. It is calculated as:
\begin{equation}
R_{\text{GP}} = 
\begin{cases} 
1 - d_{\text{norm}}(g_t^*, p_t) & \text{if } g_t^* \neq \text{null}, \\
0 & \text{if } g_t^* = \text{null},
\end{cases}
\label{eq:reward_gp} 
\end{equation}
where \( d_{\text{norm}} \) is the normalized Euclidean distance between the gripper position \( p_t \) and the inferred goal \( g_t^* \) at time \( t \).
$R_{\text{IA}}$ (Intention Alignment) represents external feedback, analogous to the activation of the mesolimbic dopamine system in response to external reward \cite{schultz2015neuronal}. This component encourages actions that align with user input, which is defined as:
\begin{equation}
R_{\text{IA}} = 
\begin{cases} 
1 & \text{if } a_t = u_t, \\ 
0 & \text{otherwise}. 
\end{cases}
\label{eq:reward_ia} 
\end{equation} \\
$R_{\text{TC}}$  (Task Completion) signifies another external reward, similar to the dopamine surge associated with goal achievement \cite{wise2004dopamine}. This provides a strong positive signal upon successful task execution and a negative signal for failure:

\begin{equation}
R_{\text{TC}} = 
\begin{cases} 
1 & \text{if task successful,} \\ 
-1 & \text{if task failed,} \\ 
0 & \text{otherwise.} 
\end{cases}
\label{eq:reward_tc}
\end{equation} \\
The weights \( \alpha \), \( \beta \), and \( \delta \) allow for tuning the relative importance of each component, analogous to individual differences in reward sensitivity \cite{montague2004computational}.

Finally, the optimal policy $\pi^*$ is learned by maximizing the expected return using Q-learning in the latent space:

\begin{equation}
Q(z_t, a_t) \leftarrow Q(z_t, a_t) + \alpha \left( r_t + \gamma \max_{a'} Q(z_{t+1}, a') - Q(z_t, a_t) \right),
\label{eq:max_qlearning}
\end{equation} \\
where $\gamma$ is the discount factor, $r_t = R(\mathbf{s}_t^\tau, a_t)$ is the reward at time $t$, and $z_t = \Psi(g_t^*, \mathbf{s}_t^\tau, \mathbf{u}_t^\tau)$ is the latent space representation.

Our ARAS formulation introduces a model-free, latent space-based formulation for shared autonomy that dynamically infers user goals using Bayesian inference while amplifying low-DoF inputs. This formulation leverages historical state and user information over a time window $\tau$ to optimize the robot’s actions in real-time, improving user alignment and task efficiency simultaneously.


\section{Case Study: Design and Implementation}
\subsection{Task Environment Overview}
We propose to evaluate our ARAS formulation as a case study by designing a pick-and-place task environment with limited user input. In this task, the robot must infer an object goal from a set of graspable objects based on user input, move toward it, grasp it, pick it up, and then transport it to one of the bins in the workspace.  Once the object is positioned above the appropriate bin, the robot must drop the object to complete the process. The challenge lies in guiding the robot's movements efficiently using the limited inputs from the user. Note that the robot has no prior knowledge about the human’s intention. We do not make any assumptions that restrict the user's intentions to be fixed throughout the task. The user can change their intentions midway, such as deciding to place the object in a different bin after initially selecting one. The only assumption we make is that the user’s inputs evolves with their intentions. 

\begin{figure*}[!ht]
\centering
\includegraphics[width=\textwidth]{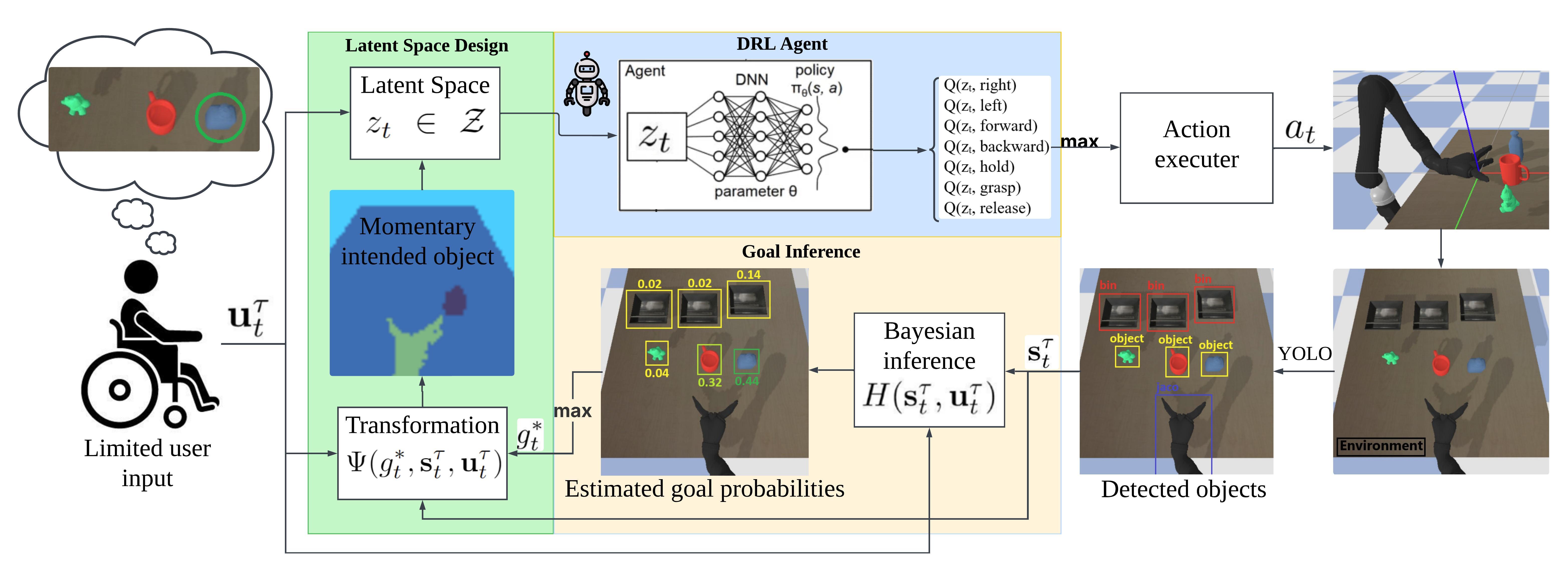}
\caption{ARAS components for a pick-and-place task case study.}
\label{Diagram}
\end{figure*}

\subsection{ARAS Components Implementation}
We project our case of study for the pick-and-place task to following proposed ARAS components, seen in Figure \ref{Diagram}:
\subsubsection{State and Action Spaces}
The state space $\mathcal{S}$ consists of both the environment state and the robot state. The environment state $\mathcal{S}_{\text{env}}$ is obtained from an RGB camera mounted above a wheelchair, as shown in Figure \ref{fig:conceptual}b, which provides a top-down view of the workspace. This camera captures the positions of the objects, bins, and the robot itself, allowing the system to continuously monitor the locations of these elements. The robot state $\mathcal{S}_{\text{robot}}$ is defined by the status of the gripper—whether it is open or closed. When the gripper is open, it signals that the robot is in a phase where it should focus on identifying and grasping objects in the scene. The open gripper prompts the system to prioritize detecting graspable objects and sets the goal distributions accordingly. Once the robot has successfully grasped an object and the gripper is closed, the robot shifts its focus to detecting bins in the environment, as the next step is to place the object in one of the designated bins. 

The action space \(\mathcal{A}\) consists of the possible movements and actions of the robotic arm, including 2D translational movements \((x, y)\), as well as grasping and releasing actions: \(\mathcal{A} = \{\text{left}, \text{right}, \text{forward}, \text{backward}, \text{hold}, \text{grasp}, \text{release}\}\), where each directional movement corresponds to a one-step displacement. The grasp action involves closing the fingers and lifting the object, while the release action opens the fingers to let go of the object. The inclusion of the "hold" action allows the robot to remain in its current position until further input is provided. This discrete action space was chosen to align with the evolving nature of human goals in shared autonomy, providing clear and interpretable feedback to the user. As the robot executes actions, the user can easily observe and adjust their inputs based on their dynamic goals. Additionally, using discrete actions simplifies the learning process by reducing the complexity of state-action mapping, enabling faster convergence during training and more stable robot behavior during execution. 

The user input, on the other hand, is highly limited, consisting of discrete signals with three possible values: \( \mathcal{U} = \{\text{left}, \text{neutral}, \text{right}\} \). This contrast highlights the need for amplification, where the robot interprets and expands the limited user inputs to select the most appropriate high-DoF action from the larger set of possible actions. The dimensionality of the user input space \( \mathcal{U} \) is represented as \( \mathbb{R}^{1 \times 3} \), while the robot's action space \( \mathcal{A} \) is represented as \( \mathbb{R}^{1 \times 7} \).

\subsubsection{Goal Space Configuration}
The goal space \( G \) in our setup is defined based on the detection of graspable objects and bins within the workspace. To accurately detect these objects, we employed a customized object detection model based on YOLOv8 \cite{ultralytics_yolo_v8} trained on our gathered dataset that includes various objects and bins relevant to the task environment. Once an object is detected, the relevant semantic information becomes part of the robot's potential goals, and the system must decide whether to move toward, grasp, or transport the object to a bin. The bins, similarly detected by the semantic model, represent the final placement locations for objects during the pick-and-place task. 

For each goal $g \in G$, we model expected behavior patterns with parameters $\mu_g$ and $\Sigma_g$, which represent the mean and variance of the expected user behavior, respectively. We initialize with a uniform prior $P(g)$ over all possible goals, reflecting an unbiased belief about the user's intentions. As interaction progresses, we collect a sliding window of $\tau=4$ time steps, capturing the system states $\mathbf{s}_t^\tau$ and user inputs $\mathbf{u}_t^\tau$. At each time step, the likelihood of the observed sequence is computed for each potential goal using the Gaussian model (Equation \ref{eq:likelihood_gaussian}), and the posterior distribution is updated using Bayes' rule (Equation \ref{eq:bayes}).  The likelihood term $P(\mathbf{s}_t^\tau, \mathbf{u}_t^\tau | g)$ is computed by aggregating likelihoods over the temporal window $\tau$ (Equation \ref{eq:likelihood}), making the inference process robust to momentary noise or user mistakes. To manage uncertainty, we introduce a confidence threshold $ \kappa =0.8$. If the maximum posterior probability falls below $\kappa$, the system infers a null goal ($g_t^* = \text{null}$) and follows a conservative action strategy, primarily following and correcting the user's input or holding its position to gather more data, rather than amplifying the input, until sufficient confidence is regained. Otherwise, the most probable goal $g_t^*$ is determined according to Equation \ref{eq:max_posterior}.

\subsubsection{Latent Space Design}
\begin{figure}[!ht]
\centering
\includegraphics[width=3.6in]{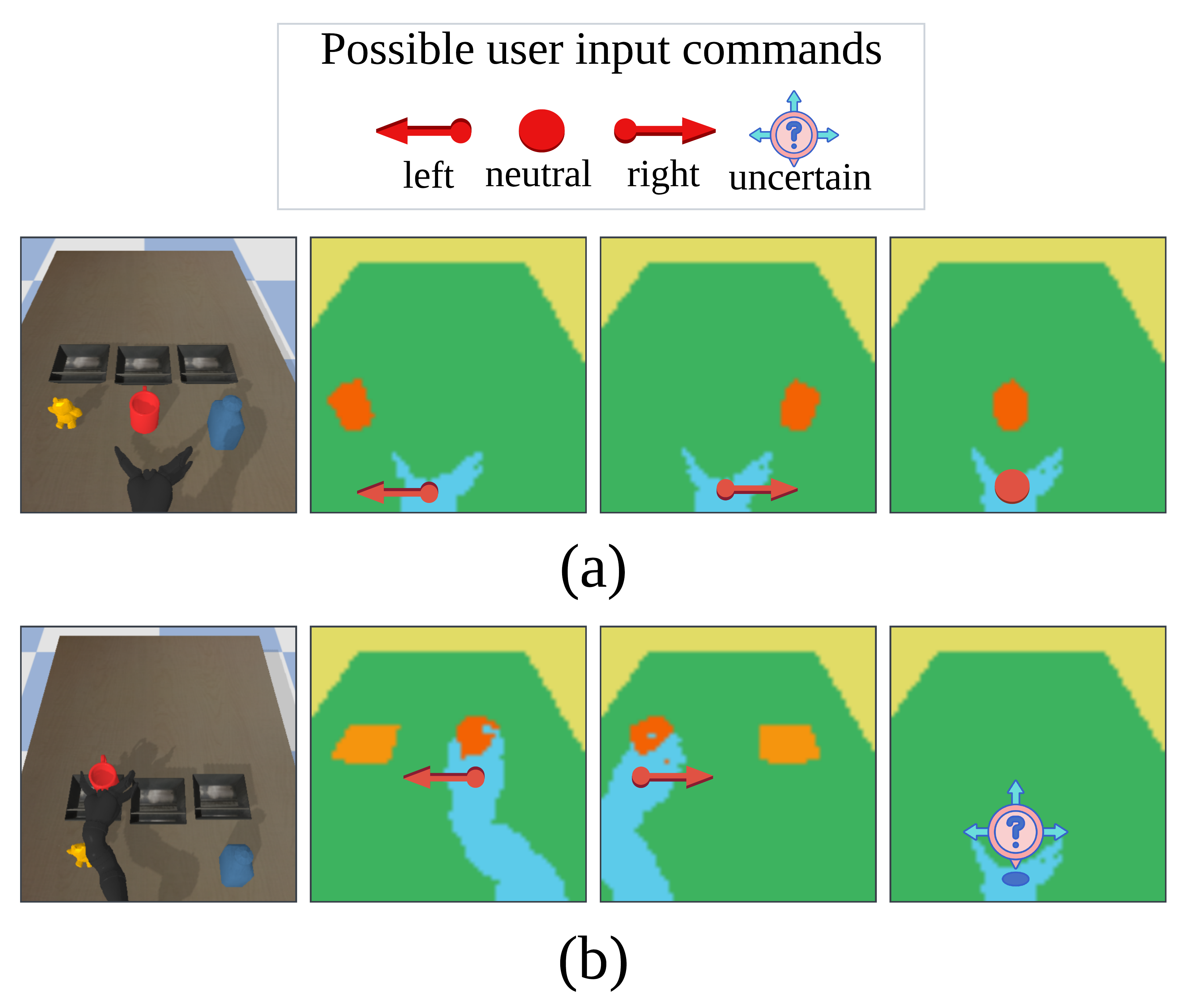}
\caption{The RGB views of the environment besides latent image representations focused on intended objects or bins based on user input commands in the (a) pick up phase and (b) drop off phase and uncertain conditions.}
\label{fig:segs_latent}
\end{figure}

The latent space \(\mathcal{Z}\) in our system is constructed by integrating limited user inputs and transformed image inputs, which are dynamically adjusted based on the estimated goal \(g_t^*\). As illustrated in Figure \ref{fig:segs_latent}, semantic segmentation techniques are employed to isolate the intended object and gripper position, effectively filtering out irrelevant information by treating it as background. This transformation is adaptive, responding to changes in the robot's state and the user's inputs to ensure task-relevant features are emphasized. When \(g_t^*\) is null, indicating uncertainty or lack of a defined goal, the system excludes all goal-related information from the latent representation. Instead, it focuses entirely on user inputs to guide action selection, thereby avoiding distractions and maintaining robust decision-making under ambiguous conditions. This adaptive behavior ensures that the system prioritizes user intent when goal inference is unclear.

The transformation from raw inputs to the latent space \(\mathcal{Z}\) is formalized through the function \(\Psi\) as below: 
\begin{equation}
    \Psi(g_t^*, \mathbf{s}_t^\tau, \mathbf{u}_t^\tau) = \phi_{\text{fusion}}(\phi_{\text{seg}}(g_t^*, \mathbf{s}_t^\tau), \mathbf{u}_t^\tau).
    \label{eq:latent_casestudy}
\end{equation} 
The function combines user input embeddings and semantic segmentation of the state and goal, seamlessly integrating these elements into a unified latent representation.


\begin{figure*}[!ht]
\centering
\includegraphics[width=\textwidth]{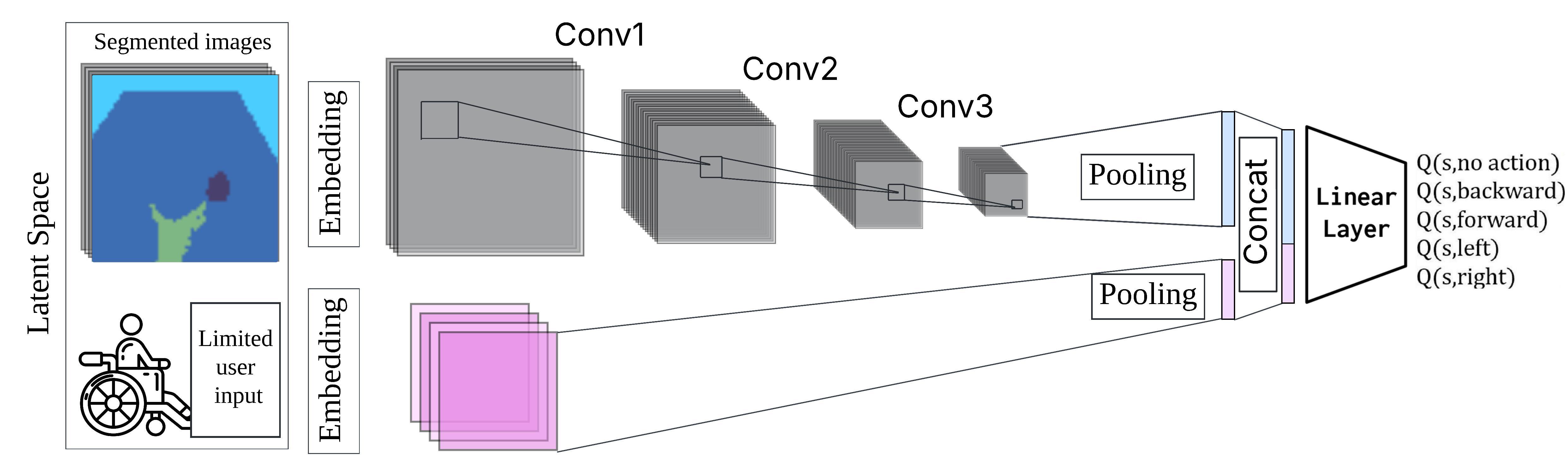}
\caption{DQN architecture for the pick-and-place task.}
\label{fig:architecture}
\end{figure*}
The policy (\(\pi\)) is implemented using a Deep Q-Network (DQN), mapping the latent space representation to robot actions. The policy determines the best action for the robot to take at each time step by considering the inferred goals, environmental constraints, and user input. The optimization process maximizes the cumulative reward \( R \) (as shown in Equation \ref{eq:reward}) using the Q-learning update rule, as formalized in Equation \ref{eq:max_qlearning}. At each time step, the Q-value \( Q(z_t, a_t) \) is updated based on the received reward and the estimated future rewards, discounted by a factor $\gamma$. This process iteratively refines the policy to select actions that maximize the expected cumulative reward, ensuring that the robot adapts effectively to evolving user inputs and task dynamics. 

\subsection{System Architecture}

\subsubsection{Network Design}

Our DQN architecture, illustrated in Figure \ref{fig:architecture}, employs a multihead design to map the latent representation \(z_t\) into discrete robot actions. The mutimodal network processes two distinct types of inputs—segmented images with dimensions \((64 \times 64)\) and user input signals \(u_t \in \{-1, 0, 1\}\)—through separate pathways, each optimized for its respective input modality. In the environmental input head, segmented images are processed through embedding layers, and then passed through a series of convolutional layers consisting of batch normalization, ReLU activation, and max pooling. This pathway progressively extracts spatial features, emphasizing task-relevant details while filtering out irrelevant information. In the user input head, the user input signals are processed separately through embedding layers, transforming them into a representation that is compatible with the features extracted from the environment input head.

The outputs from these two heads are temporally aggregated across frames and fused into a unified representation to ensure consistency over time. This integrated representation is passed through two fully connected layers: the first refines the combined features, and the second computes the Q-values corresponding to each possible robot action in the discrete action space \(\mathcal{A}\). This multihead design enables the network to independently process and integrate diverse input modalities, ensuring precise and context-aware decision-making.

\section{Simulation experiments}
\label{sec:simulation}
In this section, we elaborate on simulation experiments to train and evaluate the proposed ARAS framework in amplifying low-DoF user input into high-DoF robotic manipulation tasks. We begin by describing the training setup, including the simulation environment, task design, reinforcement learning configurations, hyperparameter tuning, and hardware specifications. Then, we explain the simulation experimental setup including task and synthetic user intention scenarios for evaluation. We leverage well-known evaluation metrics including success rates, task completion time, and cumulative rewards, to compare our system performance against state-of-the-art baseline approaches. Finally, we conducted an ablation study to investigate the individual contribution of each reward component in our reward function to the overall performance of our framework. This analysis highlights the importance of goal progress, user alignment, and task completion components in driving the robot’s behavior.

\subsection{Training setup}
Our training environment, implemented using PyBullet and OpenAI Gym, simulates a robotic arm, a table, and multiple objects and bins randomly placed on the table, providing diverse scenarios crucial for developing a robust shared autonomy system. We employed the DQN algorithm to train our reinforcement learning agent. The discrete nature of our action space aligns well with DQN's strengths in handling finite action sets, allowing for efficient learning of optimal action-value functions. For training, we leveraged our previous study \cite{rabiee2024streams} to utilize synthetic data to simulate user inputs for grasping objects and placing them in specific bins, which are randomized in each episode. As mentioned before, our approach assumes that the user input aligns with their intention, ensuring that the simulated commands accurately reflect purposeful actions directed toward achieving the task goals. Importantly, the user's goals are also randomized in each episode to introduce variability and mimic diverse task scenarios. Furthermore, in some episodes, the user's goals may change during task execution, reflecting situations where the intended outcome evolves dynamically as the task progresses.

To optimize performance, we conducted a hyperparameter search using Random Search, exploring various configurations as presented in Table \ref{tab:hyperparameters}. All configurations were trained for 20,000 episodes, with each episode terminating upon successful grasping or after reaching a maximum of 80 action steps. We employed an $\epsilon$-greedy exploration strategy, initialized at 0.9 and decaying to 0.1, with decay schedules varying across configurations. After identifying the highest-performing configuration from this search, we trained the final model for 50,000 episodes to ensure robust performance. During training, we saved the model at each peak performance and after 50,000 episodes, we finalized the best model based on the overall performance metrics. We used the Adam optimizer to train the agent's neural network, leveraging its adaptive learning rate capabilities for efficient optimization. The training was conducted on the systems specified in Table \ref{tab:hardware}, allowing for efficient parallel training and rapid iteration through hyperparameter configurations. During training, we evaluated the agent's performance using metrics such as success rate, average episode length, and cumulative reward, monitored at regular intervals to track learning progress and identify the best-performing models. The training was considered complete after 50,000 episodes, with performance plateaus monitored as a secondary convergence indicator.

\begin{table}[h!]
\centering
\caption{Hyperparameter Configurations for Random Search}
\label{tab:hyperparameters}
\begin{tabular*}{\linewidth}{@{\extracolsep{\fill}}ccc}
\toprule
\textbf{Parameter}             & \textbf{Values}                              & \textbf{Best Value} \\ 
\midrule
Learning Rate (\(\alpha\))     & 0.0005, 0.001, 0.005, 0.01                   & 0.001               \\ 
\midrule
Discount Factor (\(\gamma\))   & 0.80, 0.90, 0.95, 0.99                       & 0.95                \\ 
\midrule
Batch Size                     & 32, 64, 128, 256                             & 64                  \\ 
\midrule
Replay Buffer Size             & 10000, 50000, 100000                      & 50000              \\ 
\midrule
Target Update Freq.            & 500, 1,000, 2000                            & 1000               \\ 
\midrule
\multirow{3}{*}{\(\epsilon\)-Decay Schedule}    & 0.9 → 0.1 over 5000 steps,                  & 0.9 → 0.1 over     \\ 
                               & 0.9 → 0.1 over 10000 steps,                 & 20000 steps       \\ 
                               & 0.9 → 0.1 over 20000 steps                  &                    \\ 
\midrule
Stack Size                     & 4, 6, 8, 10                                  & 4                   \\ 
\bottomrule
\end{tabular*}
\end{table}

\begin{table}[h!]
\centering
\caption{Hardware Configurations for Training}
\label{tab:hardware}
\begin{tabular*}{\linewidth}{@{\extracolsep{\fill}}lll}
\toprule
\textbf{Computer System} & \textbf{CPU}                        & \textbf{GPU}                  \\ 
\midrule
System 1                 & Intel Xeon Silver 4214R            & NVIDIA RTX A4500              \\ 
System 2                 & AMD Ryzen 9 7950X                  & NVIDIA RTX 4090       \\ 
System 3                 & 13th Gen Intel Core i9      & NVIDIA RTX 4090       \\ 
\bottomrule
\end{tabular*}
\end{table}

\subsection{Simulation experimental setup and baselines}
The pick-and-place task is one of the fundamental activities essential for independent living and daily functionality. In our case study, we focused on this task with the aim of grasping objects requiring power grip, such as mugs. Specifically, the environment consists of three mugs and three bins (bins) randomly placed within a simulated workspace. The task is divided into pick-up and drop-off phases. In the pick-up phase, the robot must approach one of the mugs, grasp it securely, and lift it off the surface. In the drop-off phase, after successfully grasping the mug, the robot must move toward one of the bins and place the mug inside it. Failure to complete either phase, such as an unsuccessful grasp or dropping the mug outside the designated bin, results in a failed task. We consider three distinct scenarios to test the system's adaptability to dynamic user intentions:
\begin{enumerate}
    \item \textit{Fixed User Intention:} The user's goal remains unchanged throughout the task execution.
    \item \textit{Dynamic User Intention in the Pick-Up Phase:} The user intention changes randomly from one mug to another during the pick-up phase. This change occurs between 3 to 5 seconds after the phase begins.
    \item \textit{Dynamic User Intention in the Drop-Off Phase:} The user intention changes randomly from one bin to another during the drop-off phase. The change also occurs 3 to 5 seconds after the phase begins.
    \item \textit{Dynamic User Intention in Both Phases:} The user intention changes during both the pick-up and drop-off phases, introducing the highest level of uncertainty.
\end{enumerate} 
These scenarios are designed to test the system’s ability to adapt to changing goals and maintain task success under unpredictable conditions. We compared our proposed framework with two SOTA baseline methods to highlight its effectiveness:

\begin{enumerate}
    \item \textit{Regular DQN:} A standard deep Q-learning algorithm using raw image inputs and user inputs without any transformation into the latent space. This baseline tests the impact of the latent space representation in our framework.
    \item \textit{Hindsight Optimization for Action Selection (HO)} \cite{javdani2018shared}: A SOTA method for shared autonomy, which optimizes actions by retrospectively identifying the user's most likely intention at each step. 
\end{enumerate} 
These baselines provide contrasting approaches for shared autonomy and enable a thorough evaluation of our method’s strengths. To assess the performance of our system and baselines, we use the following evaluation metrics:
\begin{enumerate}
    \item \textit{Success rate:} The percentage of tasks completed successfully without failure in either the pick-up or drop-off phases.
    
    \item \textit{Task completion time:} The average time taken to complete the task successfully. A lower task completion time indicates higher efficiency in achieving the goal.
    
    \item \textit{Error actions:} The number of actions taken by the robot that result in the gripper moving farther away from the momentary intended goal, indicating inefficiencies in trajectory planning.
    
    \item \textit{Amplified actions:} The total number of actions taken by the robot that the user could not explicitly command through direct inputs. In this task, these actions include \textit{forward}, \textit{backward}, \textit{grasp}, and \textit{release}, which the robot must autonomously execute to complete the task.
    
    \item \textit{Cumulative reward:} The total reward accumulated during the task, which reflects a combination of task efficiency and adherence to the user’s goals. Higher cumulative rewards indicate better alignment with user intentions and task success.
    
    \item \textit{Total user inputs:} The total number of user inputs recorded during the task. This is calculated as the absolute sum of all user inputs provided at every 0.05-second interval, representing the user's involvement in the task.
\end{enumerate}

Furthermore, to evaluate the importance of each component in the reward function, we performed an ablation study by selectively removing individual terms and analyzing their impact on task performance. The reward function, as defined in Equation \ref{eq:reward} and described in Section \ref{sec:formulation}, consists of three components: goal progress (\(R_{GP}\)), user alignment (\(R_{IA}\)), and task completion (\(R_{TC}\)). For this analysis, we trained the model with one term removed at a time and observed changes in cumulative reward and task completion time. 

\begin{table*}[h!]
\centering
\caption{Comparison of Simulation Results Across Methods and Scenarios}
\label{tab:simulation_results}
\begin{tabular*}{\textwidth}{@{\extracolsep{\fill}}lcccccc}
\toprule
\textbf{Scenario}                              & \textbf{Method}     & \textbf{Completion Time (s)} & \textbf{Success Rate (\%)} & \textbf{Total Inputs} & \textbf{Error Actions (s)} & \textbf{Amplified Actions (s)} \\ 
\midrule
\multirow{3}{*}{Fixed User Intention}          & HO                 & 36.03                        & 85.80                       & 360.72                   & 6.46                      & 18.24                         \\ 
                                               & DQN                  & 35.62                         &88.44                       & 292.33                   & 5.44                      & 19.64                         \\ 
                                               & ARAS                & 31.48                        & 99.76                       & 274.45                   & 4.24                      & 20.34                         \\ 
\midrule
\multirow{3}{*}{Dynamic Intention in Pick-Up}  & HO                 &39.12                          & 65.80                       & 378.20                   & 6.90                     & 18.52                         \\ 
                                               & DQN                  & 36.30                          & 95.40                       & 323.75                   & 5.36                      & 20.15                         \\ 
                                               & ARAS                & 35.13                         & 96.55                       &325.28                   & 4.70                      & 19.86                         \\ 
\midrule
\multirow{3}{*}{Dynamic Intention in Drop-Off} & HO                 & 38.68                         & 84.80                       & 387.44                   & 6.22                     & 18.23                         \\ 
                                               & DQN                  &36.44                          & 89.4                       & 296.54                   & 4.80                      & 21.37                         \\ 
                                               & ARAS                & 33.24                         & 99.44                       & 276.24                   & 4.14                      & 21.56                         \\ 
\midrule
\multirow{3}{*}{Dynamic Intention in Both}     & HO                 & 39.56                         & 69.40                       & 417.87                   & 6.78                     & 18.49                         \\ 
                                               & DQN                  & 38.74                         & 95.80                       & 343.24                   & 5.03                     & 19.54                         \\ 
                                               & ARAS                & 37.80                         & 97.76                       &342.14                   & 4.50                      & 22.68                         \\ 
\bottomrule
\end{tabular*}
\end{table*}

\begin{figure*}[!ht]
\centering
\includegraphics[width=\textwidth]{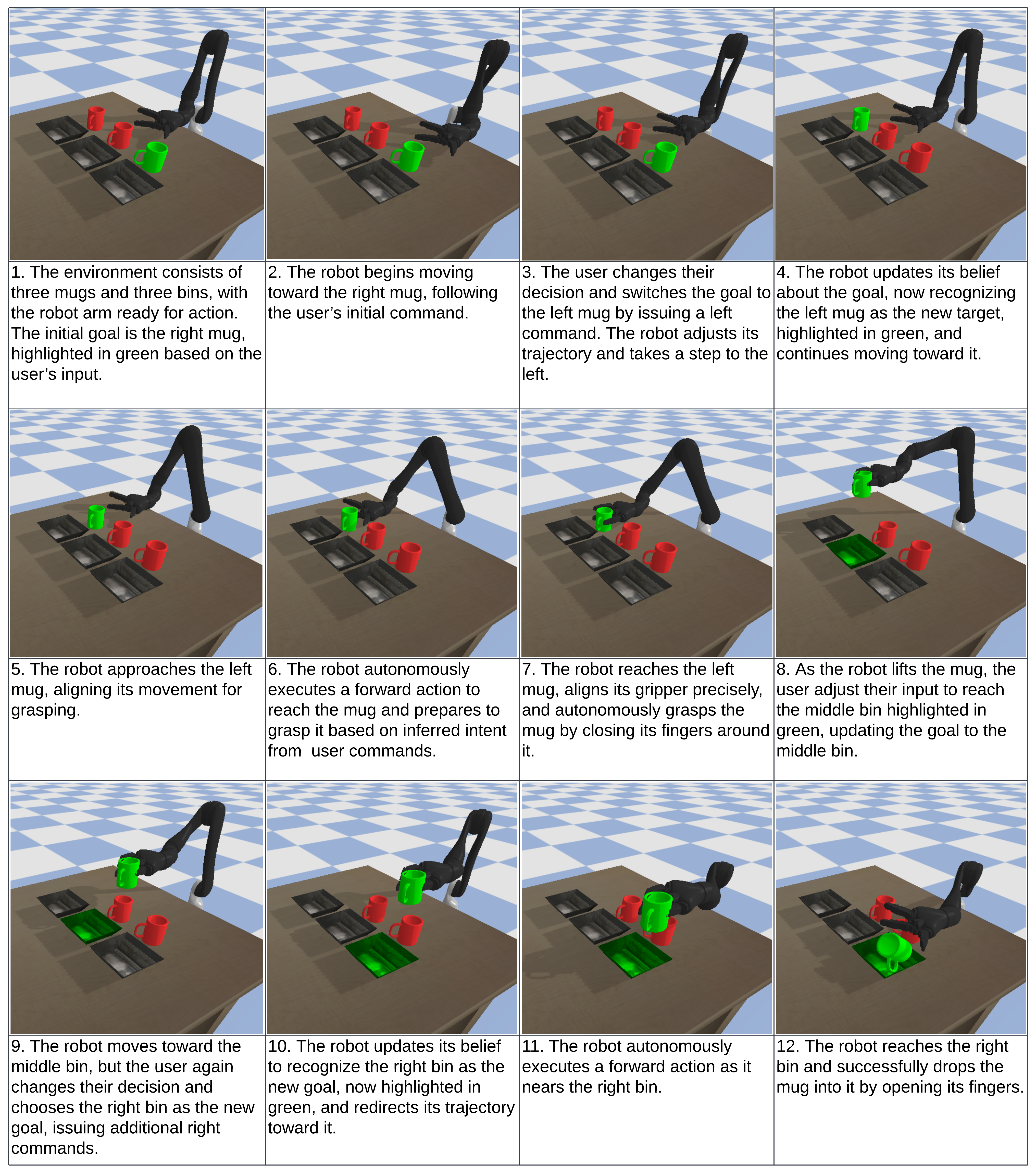}
\caption{An example of ARAS adapting to dynamic user intentions during both the pick-up and drop-off phases in a shared autonomy framework.}
\label{fig:evolve_sequence}
\end{figure*}

\subsection{Results and analysis}

The simulation results in Table \ref{tab:simulation_results} reveal clear performance differences among ARAS, DQN, and HO across progressively more challenging intention scenarios. In the fixed‑intention condition, where the user’s goal remains constant, all three policies complete the task reliably, but ARAS leads with the shortest mean completion time (31.48s) and an almost perfect success rate (99.76\%), while also demanding the fewest total inputs and exhibiting the lowest error action time. DQN follows as a strong second, surpassing HO in success rate (88.44\% vs. 85.80\%) and reducing both input burden and error actions.

When the user’s goal changes once during the pick‑up phase, adaptability becomes critical. ARAS sustains a high success rate (96.55\%) and keeps completion time near 35s, indicating rapid realignment to the new target with minimal user effort. DQN also adapts effectively, achieving 95.40\% success at a slightly higher error action cost, whereas HO’s performance drops markedly to 65.80\%, accompanied by the greatest input load and error action duration in this scenario. A goal switch during the drop‑off phase produces a similar pattern. ARAS again excels, reaching 99.44\% success and finishing in 33.24s while keeping error actions to 4.14s. DQN maintains a respectable 89.40\% success rate, outperforming HO (84.80\%), but it still incurs longer completion times than ARAS. HO remains the least efficient, reflecting its limited capacity to re‑plan late in the task.

The most demanding case, goal changes in both pick‑up and drop‑off phases, accentuates these trends. ARAS preserves top performance, completing the task in 37.80s with a 97.76\% success rate and the lowest error‑action total (4.50s). DQN remains resilient, achieving 95.80\% success yet requiring more user inputs than ARAS. HO struggles most in this double‑switch condition, with success dropping to 69.40\% and all user‑effort metrics rising sharply.

Across all scenarios, total inputs, error action time, and amplified action time rise with increasing intention complexity. Nevertheless, ARAS consistently moderates these increases better than HO and, in most cases, more effectively than DQN. These findings underscore the advantages of dynamic goal alignment and action amplification: ARAS’s latent‑space policy provides the greatest robustness to shifting user goals, DQN delivers competitive success rates, though at higher user cost, and HO’s heuristic approach shows clear limitations as goal dynamics become more complex.

Figure \ref{fig:evolve_sequence} illustrates an example of ARAS handling dynamic user intentions during both the pick-up and drop-off phases in a shared autonomy framework. The task involves a robot arm interacting with three mugs and three bins in a simulated environment. This example highlights ARAS’s capability to dynamically update its goal belief and adapt its actions in response to evolving, limited user inputs, while autonomously amplifying essential actions such as moving forward, grasping, and dropping. The figure showcases how ARAS integrates user guidance with autonomous decision-making to achieve reliable task execution, even under conditions of uncertainty and shifting goals.

\begin{figure}[!ht]
\centering
\includegraphics[width=3.6in]{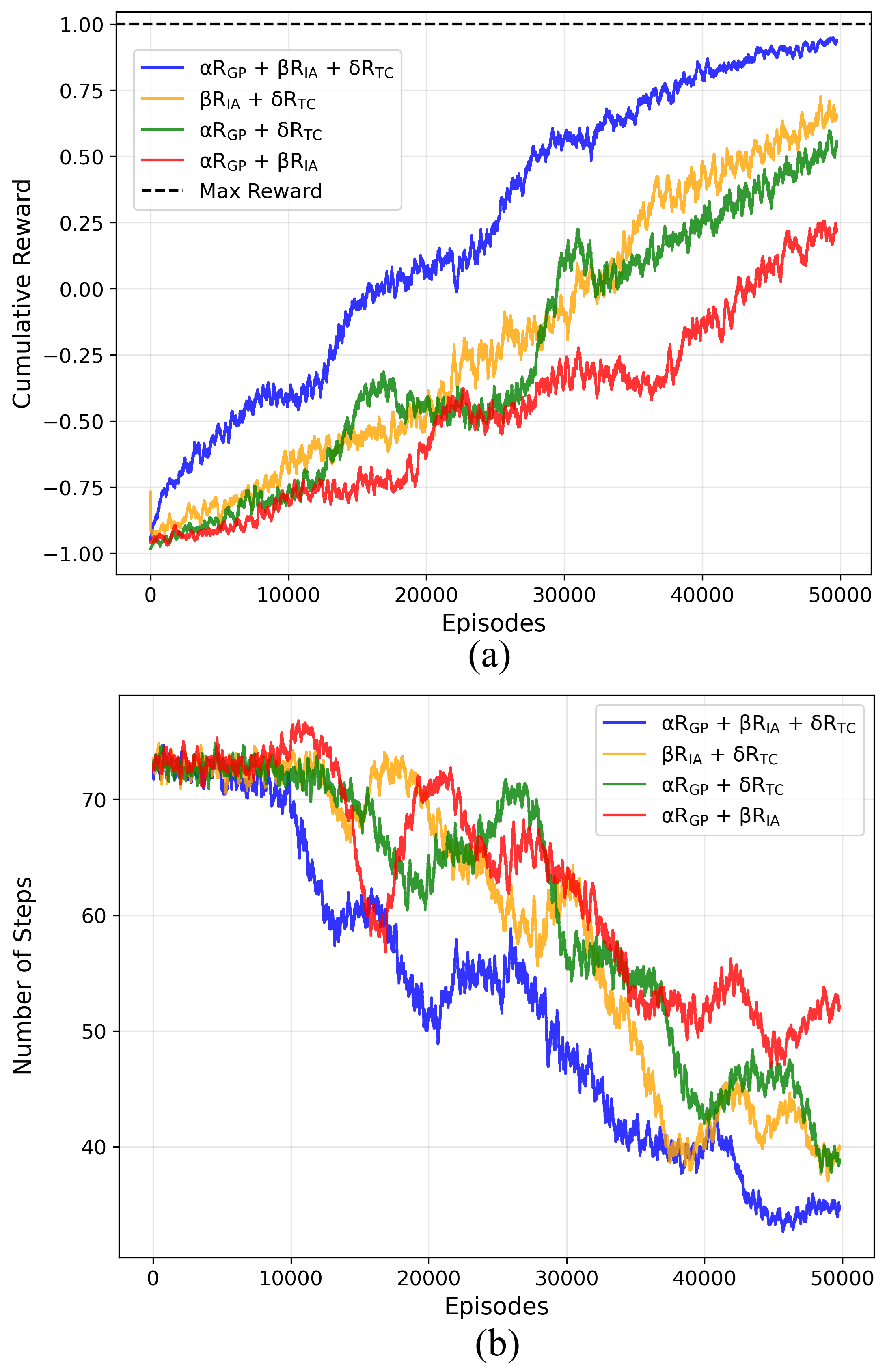}
\caption{Visualization of the ablation study results for different reward function configurations over 50,000 training episodes. (a) Cumulative reward trends for various configurations, showing the learning progression. (b) Number of steps required to complete tasks across training episodes, comparing efficiency for each reward configuration. This figure demonstrates the impact of individual reward terms on learning dynamics and task execution.}
\label{fig:reward_trends}
\end{figure}

Figure \ref{fig:reward_trends} illustrates the results of the ablation studies conducted on different configurations of the reward function, focusing on cumulative rewards and the number of steps during training over 50,000 episodes. The findings highlight the impact of each reward term on the system's ability to learn efficient task execution. In the cumulative reward plot (Figure \ref{fig:reward_trends}a), the configuration using all three reward terms (\(\alpha R_{GP} + \beta R_{IA} + \delta R_{TC}\)) exhibits the best performance, with the curve converging the fastest and reaching the highest cumulative reward. This result demonstrates the effectiveness of combining goal progress (\(R_{GP}\)), intention alignment (\(R_{IA}\)), and task completion (\(R_{TC}\)) terms to guide the agent’s learning process. The other configurations achieve lower cumulative rewards, reflecting the influence of the excluded terms. For example, the configuration excluding \(R_{GP}\) (\(\beta R_{IA} + \delta R_{TC}\)) achieved relatively high rewards but does not match the performance of the full reward function, indicating that \(R_{GP}\) contributes notably to improving task efficiency. Similarly, removing \(R_{IA}\) (\(\alpha R_{GP} + \delta R_{TC}\)) results in lower cumulative rewards, suggesting that \(R_{IA}\) is crucial for aligning the agent’s actions with user inputs. The configuration without \(R_{TC}\) (\(\alpha R_{GP} + \beta R_{IA}\)) shows the slowest convergence and the lowest cumulative rewards, highlighting the critical role of \(R_{TC}\) in ensuring successful task completion. This result is unsurprising, as \(R_{TC}\) represents the largest reward or penalty in the framework, directly tied to whether the task is completed successfully or not. Its absence leaves the agent without a strong incentive to prioritize task completion, resulting in inefficient learning and significantly lower performance compared to configurations that include this term. This emphasizes that \(R_{TC}\) is the most essential component of the reward function, driving the overall success of the task.

In the number of steps plot (Figure \ref{fig:reward_trends}b), the configuration using all three reward terms again outperforms the others, achieving the fewest steps required to complete the task. This result aligns with the cumulative reward trends, as higher rewards often correlate with increased efficiency. The other configurations show varying levels of inefficiency, with the configuration excluding \(R_{TC}\) requiring the highest number of steps throughout training. This indicates that \(R_{TC}\) plays a critical role in driving the agent to complete tasks effectively. Configurations excluding \(R_{GP}\) or \(R_{IA}\) exhibit intermediate performance, further emphasizing the contributions of these terms to learning efficiency. Overall, the results demonstrate that incorporating all three reward terms leads to the most efficient and reliable learning process. Each term contributes uniquely to the overall performance: \(R_{GP}\) ensures efficient movement toward goals, \(R_{IA}\) aligns actions with user intentions, and \(R_{TC}\) motivates task completion. 

\section{User study with the physical robot}

In this section, we extend our evaluation of the ARAS framework to a real user study, comparing its performance against the two ends of a shared control spectrum known as completely manual control and fully autonomous systems. While manual control represents the extreme of complete user agency, requiring precise input for every action, fully autonomous systems eliminate user involvement. Users only intervened to select the target object at the start of each trial. These two extremes provide critical baselines to evaluate the strengths and weaknesses of ARAS, which aims to strike a balance by amplifying limited user inputs and autonomously adapting to evolving goals. A total of 23 healthy participants (X female, Y male; age range 20–40) were recruited for the study. All participants provided written informed consent prior to participation. The experimental procedure was reviewed and approved by the Institutional Review Board (IRB), ensuring compliance with ethical standards for research involving human subjects.

\subsection{Hypotheses}
We tested the following hypotheses in our experiments: \\
H1 ARAS will enable faster task completion times compared to manual control. \\
H2 ARAS will significantly reduce the effort required from users compared to manual control. \\
H3 Users will feel more in control and flexible with ARAS than with an autonomous system while experiencing less operational burden than with manual control. \\
H4 Users will report higher satisfaction and confidence when using ARAS compared to both manual control and fully autonomous systems.

\subsection{User study experimental setup}
\begin{figure}[!ht]
\centering
\includegraphics[width=3.5in]{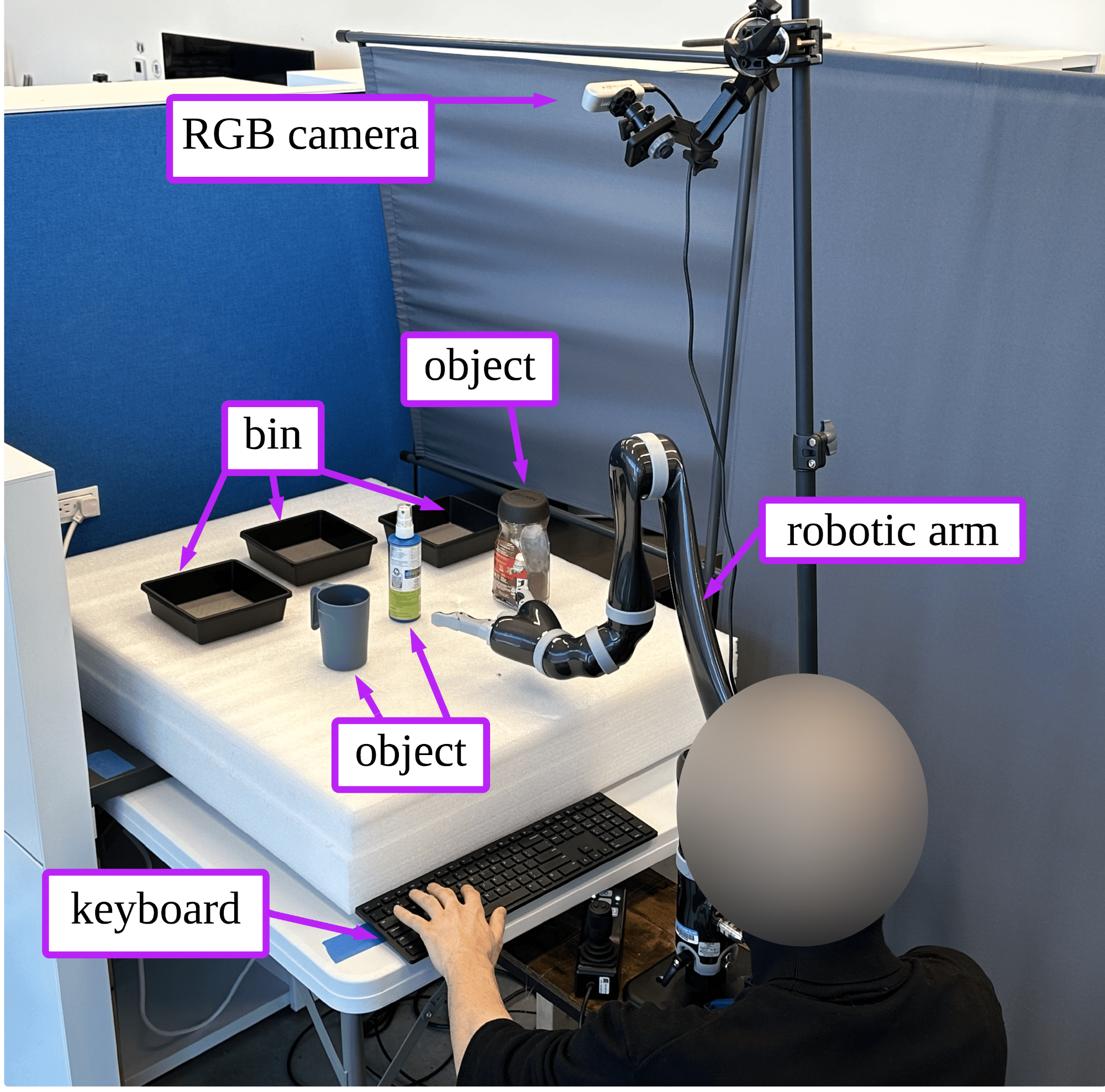}
\caption{Experimental setup for the user study, showing a participant controlling the robotic arm with a keyboard while an overhead RGB camera captures the workspace. Multiple objects and bins are placed on a foam platform to facilitate pick-and-place tasks, demonstrating the key components (robotic arm, camera, keyboard, and objects) used to evaluate the system’s performance.}
\label{fig:env_comp}
\end{figure}

The experimental setup involved a pick-and-place task similar to the one used in the simulation, with three objects and three bins placed in the same positions across all trials (Figure \ref{fig:env_comp}. An Intel RealSense D435 camera was mounted overhead to provide a top-down view of the environment. The robotic system used for the experiment was a Kinova JACO2 arm equipped with a three-finger gripper. The robot was mounted on a small table set at a height similar to that of a wheelchair to closely mimic real-world conditions. Participants teleoperated the robotic arm using a wireless keyboard, with keys mapped to actions such as left, right, forward, backward, grasp, and release. This setup allowed users to directly control the robot's movements and interactions during the manual control and ARAS modes. In the fully autonomous mode, participants provided high-level voice commands through a headphone microphone to express their intentions. These commands were interpreted to guide the robot's actions, but a hidden professional operator teleoperated the robot to perform the tasks with near-perfect precision. This approach was designed to simulate an ideal fully autonomous system, creating an ideal baseline that neither manual nor ARAS modes could surpass in terms of accuracy. In our proposed user study, we built a structured framework for consistency among participants and later statistical analysis within our population study. Supplementary materials extended the results to different scenarios and unstructured setups.  

\begin{figure*}[!ht]
\centering
\includegraphics[width=7in]{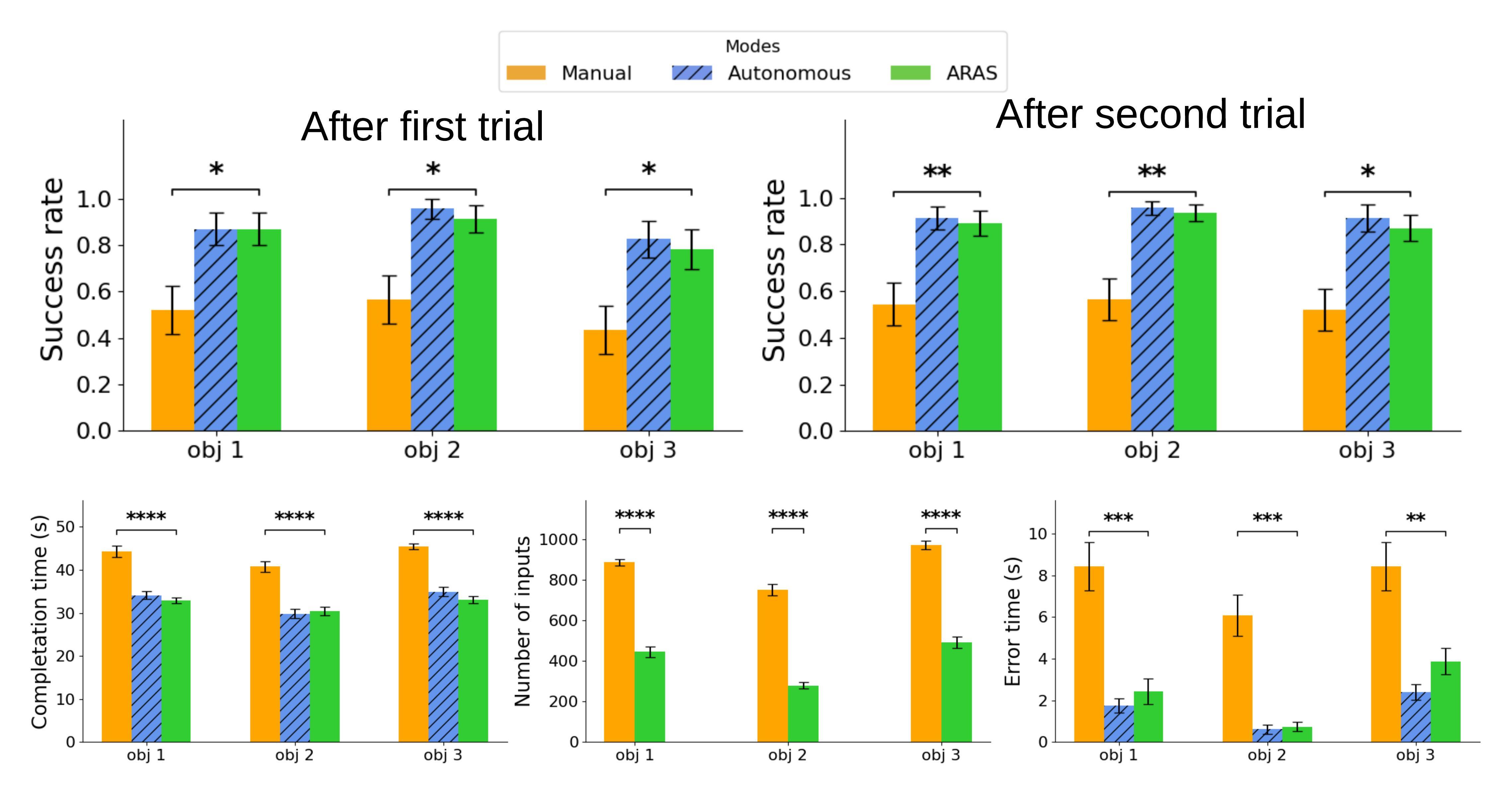}
\caption{Comparison of objective measures across manual, ARAS, and autonomous modes for three objects (obj 1: left cup, obj 2: middle cup, obj 3: right cup). The autonomous mode, illustrated by the hatched blue bars, is considered the ideal scenario, theoretically not surpassable in precision. Metrics include the average success rates across participants after the first and second trials, along with the average task completion times, inputs required, and error times, calculated across all trials and participants. Statistically significant differences are indicated by stars (* \(p<0.05\), ** \(p<0.01\), *** \(p<0.001\), **** \(p<0.0001\)).}
\label{fig:objective}
\end{figure*}

\subsection{Procedure}

Participants performed a total of 90 trials, divided into 30 trials for each of the three modes: manual, autonomous, and ARAS. Each mode consisted of three tasks, involving pick-and-place actions for three cups and their corresponding bins. The cups were placed on the left, middle, and right, with each task requiring the participants to move the robotic arm, grasp the cup, and place it in the corresponding bin. The tasks were designed to reflect varying levels of difficulty. The first task, involving the left cup and bin, required moderate alignment effort. The second task, with the middle cup and bin, was the easiest due to its alignment with the gripper’s natural position. The third task, involving the right cup and bin, was the hardest, as it required the robotic arm to move farther from its initial position, increasing task complexity and time. Each task was repeated 10 times per mode, resulting in 30 trials per mode and a total of 90 trials per participant.

Participants used a wireless keyboard to control the robotic arm during the trials. In the manual mode, participants had to perform all actions themselves, including aligning the gripper with the cup and bin, grasping, and releasing the cup. In the ARAS mode, participants only needed to input directional commands (left and right) to guide the robot, which autonomously amplified and executed the remaining actions, such as moving forward, grasping, and dropping. In the autonomous mode, participants provided high-level voice commands via microphones, and the robotic arm operated by a hidden professional for execution. The order of the modes was randomized to mitigate order effects.  Before the trials, participants received a user sheet containing detailed instructions, task descriptions, and an explanation of the controller inputs. The examiner provided an overview of the procedure and clarified any questions before obtaining the participant’s consent to proceed. After the trials, participants were asked to complete a 5-point Likert scale questionnaire designed to evaluate subjective measures for each mode. The questionnaire assessed aspects such as:

\begin{itemize}
    \item \textit{Satisfaction:} How satisfied participants were with each mode.
    \item \textit{Effort:} The perceived effort required to complete the tasks (higher scores indicated less effort).
    \item \textit{Confidence and Trust:} How much participants trusted the robot and anticipated its next action.
    \item \textit{Flexibility:} The ability of the system to adapt to changing user intentions.
    \item \textit{Future Use:} How likely participants were to use each system in the future.
    \item \textit{Level of Assistance Preference:} How much assistance participants preferred in general.
\end{itemize}
To ensure consistency, the questionnaire was designed such that higher scores always indicated better outcomes. Additionally, the performance of each mode was evaluated using the objective measures outlined in Section \ref{sec:simulation}, such as task success rates, completion times, number of inputs, and error rates.

\subsection{Results and analysis}

The user study results are divided into three categories: objective measures, subjective measures, and trajectory efficiency. Objective measures quantify task performance, including metrics such as success rates, completion times, and error times. Subjective measures capture participant perceptions of the system, including satisfaction, effort, ease of use, and trust. Trajectory efficiency evaluates the robot's motion optimization during task execution. All data reported in this section represent the grand average across all trials and participants to ensure consistency and reliability in the analysis.

For statistical analysis, the Wilcoxon signed-rank test was used, as it is a non-parametric test suited for paired data, such as participant responses across the three modes. Statistical significance in the results is indicated using standard symbols. If \(p > 0.05\), the result is labeled as "ns" (not significant). For \(p < 0.05\), significance is denoted with "*", \(p < 0.01\) with "**", \(p < 0.001\) with "***", and \(p < 0.0001\) with "****". These annotations provide clear indications of the statistical strength of differences observed between modes.

\subsubsection{Objective Measures}

Figure \ref{fig:objective} provides evidence supporting our hypotheses by comparing the objective measures across three modes—manual, ARAS, and autonomous—evaluated over three objects corresponding to the task setup: obj 1 (left cup), obj 2 (middle cup), and obj 3 (right cup). The comparisons include success rates after one and two attempts, task completion times, the number of inputs required, and error times. Statistically significant differences between modes are indicated by the presence of stars, with more stars denoting stronger significance.

In terms of success rates, ARAS, and autonomous modes demonstrate significantly higher success rates compared to manual control across all three objects after one attempt. After two attempts, success rates improve across all modes, with ARAS maintaining performance comparable to the autonomous mode. Notably, the middle cup (obj 2) exhibits the highest success rates across all modes, likely due to its easier alignment with the gripper, while the right cup (obj 3) remains the most challenging, requiring more precise movements. Overall, the average success rates across all trials and subjects were 86.46\% for manual control, 95.60\% for autonomous mode, and 92.88\% for ARAS. The difference between ARAS and the manual mode was marginal (\(p = 0.086\)), but ARAS demonstrates higher performance and performs only slightly below a professional operator executing the task in the autonomous mode. When examining the first attempts, it is evident that manual control posed initial challenges for participants. Misclicks, such as slipping the mug during grasping or misalignment before the grasp action, frequently cause failures, indicating a potential learning curve for participants in the manual mode. In contrast, participants adapted quickly to ARAS, enabling high performance without a steep learning curve. 

Task completion times further emphasize the efficiency of ARAS, directly supporting H1. Completion times for ARAS are significantly shorter than manual control across all objects, reflecting its ability to amplify user inputs and optimize task execution. While the autonomous mode, operated by a professional, achieved slightly faster completion times, ARAS remained competitive, as expected given the ideal nature of the autonomous setup. Among the objects, the right cup (obj 3) takes the longest time in all modes due to its greater distance from the robot's starting position, while the middle cup (obj 2) is completed the fastest. ARAS also significantly reduces the number of inputs required from users compared to manual control, as shown across all objects. This reduction in input effort aligns with H2, highlighting ARAS’s ability to autonomously handle complex actions, such as grasping and dropping while minimizing the operational load on the user. 

Finally, ARAS and autonomous modes exhibit significantly lower error times compared to manual control, reflecting better task accuracy and fewer mistakes. Error times for ARAS are slightly higher than those for the autonomous mode, but the differences are minor, demonstrating ARAS's ability to provide precision while involving user input. Similar to completion times, the error times for the right cup (obj 3) are the highest due to the increased task difficulty, while the middle cup (obj 2) exhibits the lowest error times.

\begin{figure*}[!ht]
\centering
\includegraphics[width=7.2in]{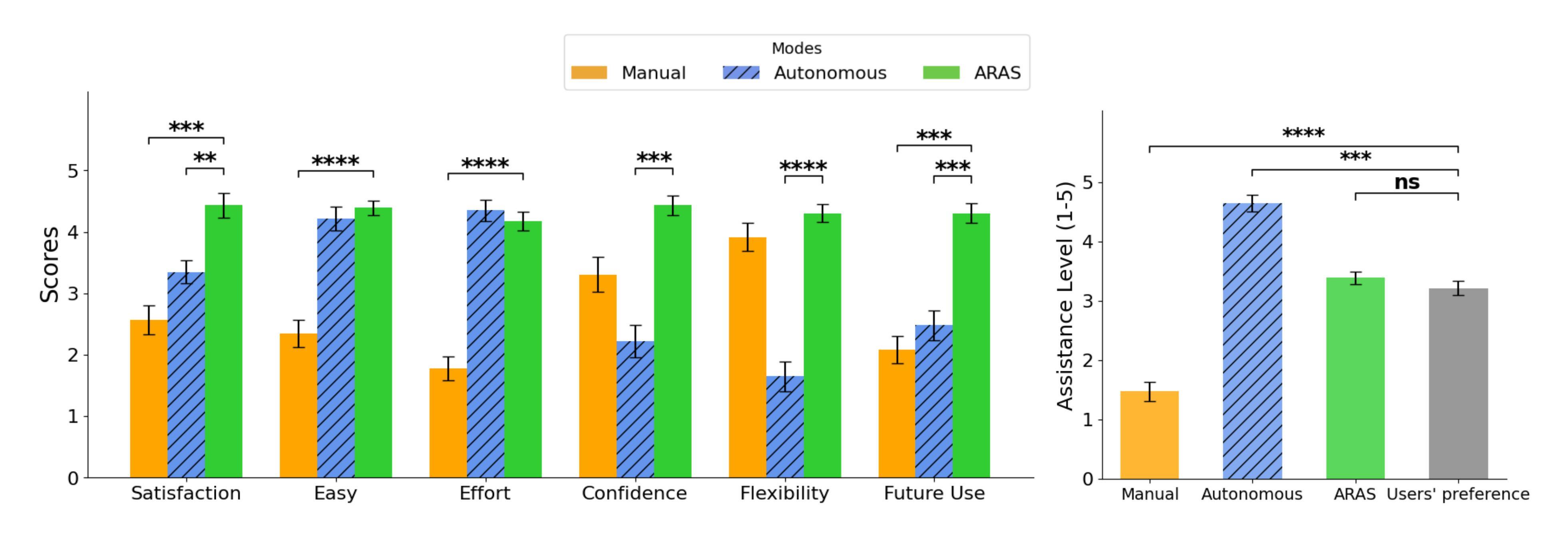}
\caption{Subjective measures collected through a 5-point Likert scale questionnaire, evaluating participant feedback on each mode. (a) The average scores for satisfaction, ease of use, effort, confidence, flexibility, and future use likelihood across the three modes. Participants' ratings of the perceived level of assistance provided by each mode and their overall preference for assistance. Statistically significant differences are indicated by stars (* \(p<0.05\), ** \(p<0.01\), *** \(p<0.001\), **** \(p<0.0001\)).}
\label{fig:subjective}
\end{figure*}

\subsubsection{Subjective Measures}
Figure \ref{fig:subjective} presents the subjective measures collected through a 5-point Likert scale questionnaire, evaluating participant feedback on the manual, autonomous, and ARAS modes. Figure \ref{fig:subjective}a shows the average scores for satisfaction, ease of use, effort, confidence, flexibility, and future use likelihood across the three modes. As it is shown, ARAS and autonomous modes received higher ratings for satisfaction, ease of use, and confidence compared to manual control. ARAS achieved particularly high scores for flexibility, reflecting its ability to adapt effectively to user input, while future use likelihood was also rated the highest for ARAS. These findings support H3, which hypothesizes that users would feel more in control and experience greater flexibility with ARAS compared to autonomous systems while reducing the operational burden seen in manual control. Effort scores indicate that both ARAS and autonomous modes required less effort from participants compared to manual control, directly supporting H2. Additionally, participants’ high confidence and satisfaction ratings for ARAS support H4, which hypothesizes that users would report higher satisfaction and confidence with ARAS compared to both manual and autonomous modes.

Figure \ref{fig:subjective}b illustrates participants' ratings of the perceived level of assistance provided by each mode and their overall preference for assistance. Participants rated the level of assistance provided by the autonomous mode as the highest, followed by ARAS, with manual control rated as the lowest. Notably, there was no significant difference between ARAS and participants' preferred level of assistance, suggesting that ARAS offers an optimal combination of autonomy and user control. This result further supports H3, indicating that ARAS aligns closely with participants' desired balance between assistance and user involvement. 

\begin{figure}[!ht]
\centering
\includegraphics[width=3.5in]{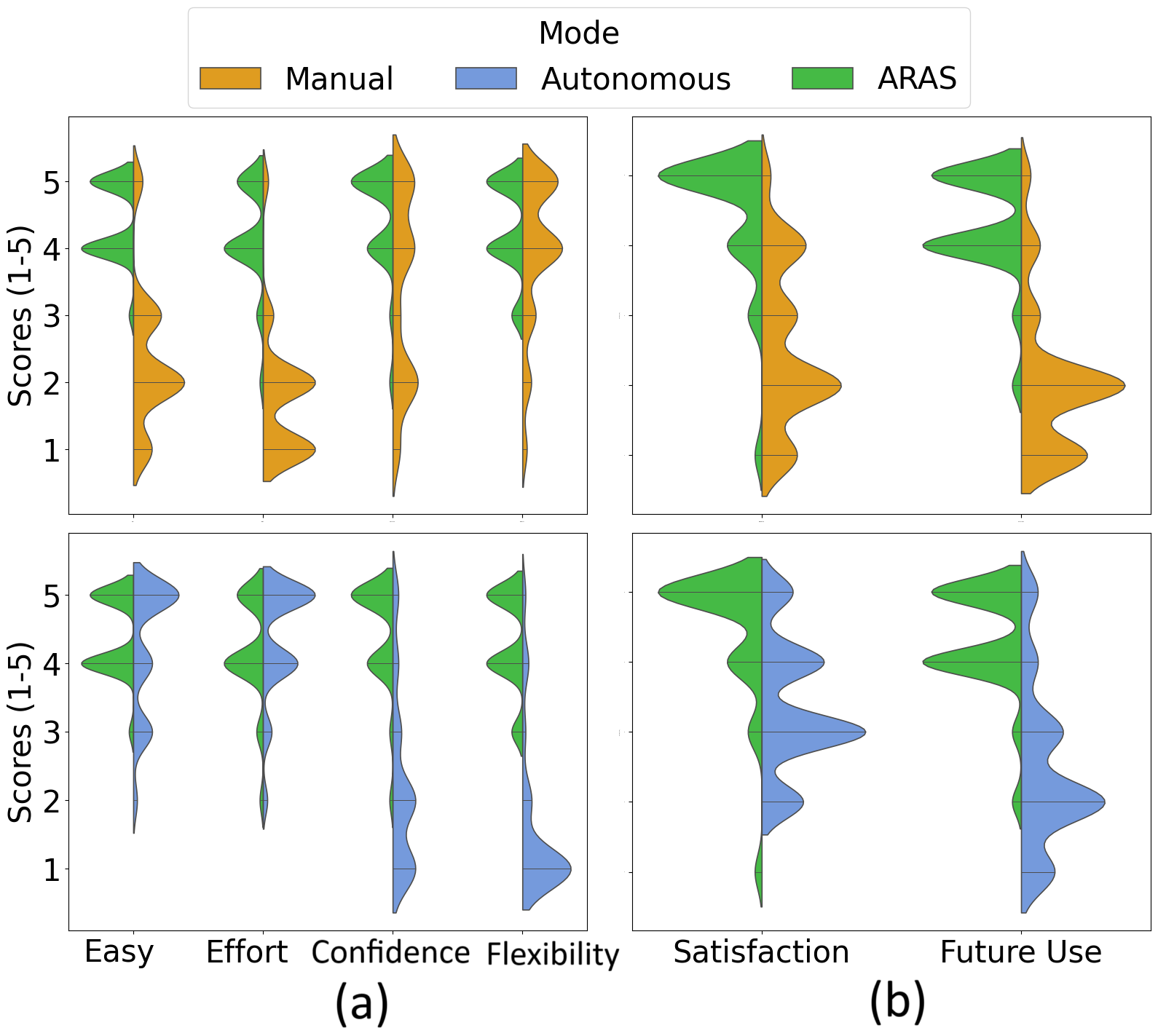}
\caption{Violin plots showing the distribution of subjective measure scores (1–5) for each mode. (a) compares ease of use, effort, confidence, and flexibility. (b) illustrates satisfaction and future use likelihood}
\label{fig:violin}
\end{figure}

Figure \ref{fig:violin} presents violin plots comparing subjective measures—ease of use, effort, confidence, flexibility, satisfaction, and future use likelihood—across the three modes: manual, autonomous, and ARAS. The plots show the distribution of participant scores (1–5), with higher concentrations toward the upper end indicating better performance. In Figure \ref{fig:violin}a, ARAS exhibits higher scores for ease of use and lower effort compared to manual control, with distributions concentrated near the upper range, supporting H2. Manual control shows broader distributions, reflecting greater variability and more effort. While ARAS and autonomous modes are similar in ease of use and effort, ARAS demonstrates better performance in flexibility and confidence, with distributions more concentrated in the upper range. High flexibility scores for ARAS support H3, indicating its adaptability to user input.

In Figure \ref{fig:violin}b, ARAS achieved higher distributions for satisfaction and future use likelihood, with scores concentrated at the upper end, outperforming manual control and autonomous mode. These results align with H4, showing participants’ higher satisfaction and a greater likelihood of choosing ARAS for future use.

\begin{figure}[!ht]
\centering
\includegraphics[width=3.5in]{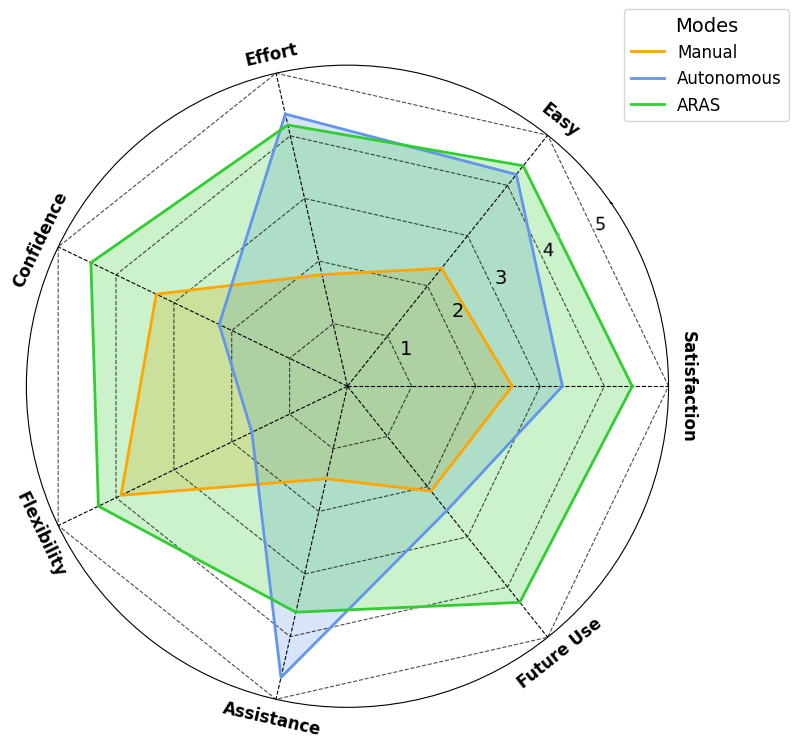}
\caption{Spider plot comparing subjective measures across manual, autonomous, and ARAS modes}
\label{fig:spider}
\end{figure}

Figure \ref{fig:spider} highlights the balance achieved by ARAS across subjective measures compared to manual and autonomous modes. While the autonomous mode excels at the assistance level, ARAS offers a more even distribution across metrics, blending user control and support. Despite requiring slightly higher effort than the autonomous system, ARAS is preferred by participants, as reflected in its higher ratings for satisfaction and future use. This preference suggests that users value the balance ARAS provides, maintaining a sense of control and flexibility while still reducing effort. By bridging the gap between manual control and full autonomy, ARAS enables user engagement without sacrificing assistance, aligning closely with participant preferences for a shared autonomy system that effectively integrates control and support.

\subsubsection{Trajectories Analysis}
This section examines the trajectories generated by the ARAS during the pick-and-place task. Trajectory analysis provides insights into the efficiency, smoothness, and goal-directed behavior of ARAS, highlighting differences in how this system executes actions to complete the task. By analyzing trajectory paths, we aim to assess the ability of ARAS to amplify user inputs while maintaining efficient and adaptive motion

\begin{figure*}[!ht]
\centering
\includegraphics[width=7.2in]{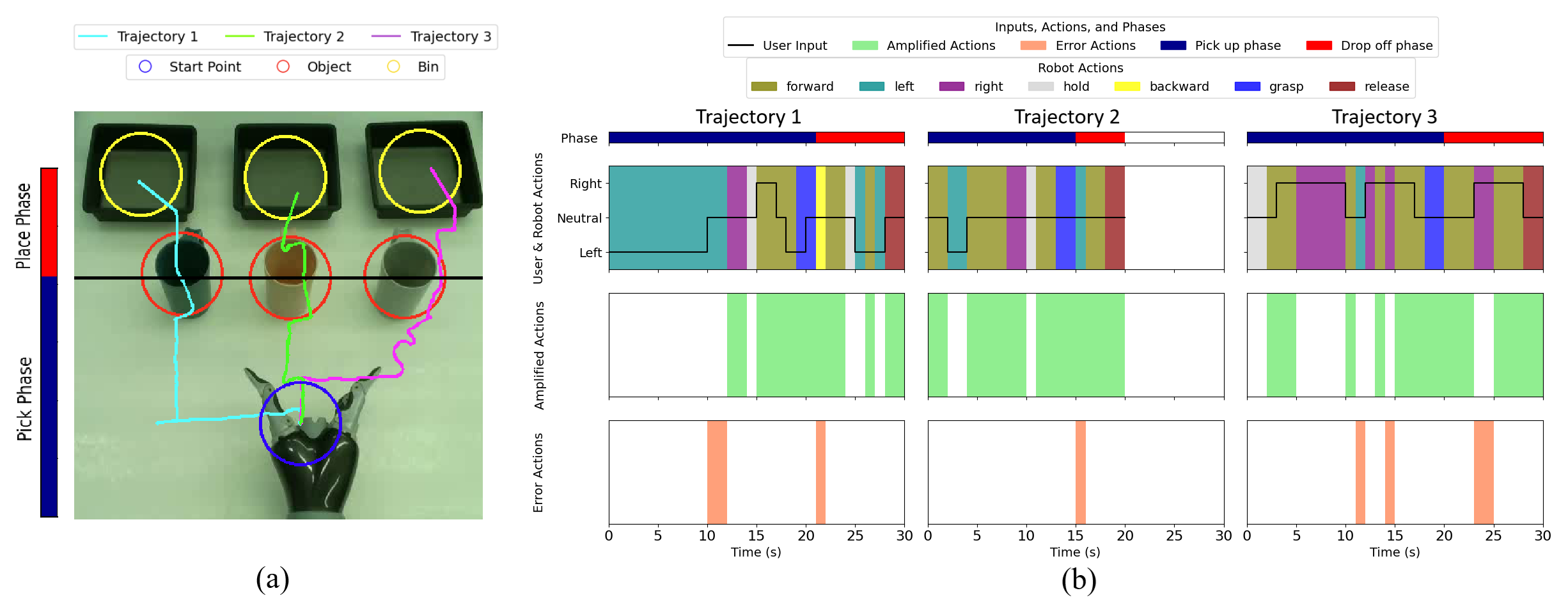}
\caption{Example of trajectories generated using the ARAS framework. (a) illustrates the spatial trajectories for three tasks, showing the robot's path from the start point to the pick-up and drop-off locations. (b) presents the temporal dynamics of user inputs, robot actions, amplified actions, and error actions for each trajectory.}
\label{fig:trajfull}
\end{figure*}

Figure \ref{fig:trajfull} provides an example of trajectories executed using the ARAS framework, illustrating both the spatial paths of the robot and the temporal dynamics of user and robot actions. Figure \ref{fig:trajfull}a shows the overall trajectories for three different tasks, where the robot begins at a designated start point, picks up objects from their locations, and places them in corresponding bins. The trajectories are color-coded to differentiate between the three tasks, and distinct markers indicate the start points, object positions, and bin locations. The plot visually separates the pick-up and drop-off phases, highlighting the path adjustments made by the robot in response to user inputs and amplified actions.

Figure \ref{fig:trajfull}b delves deeper into the temporal aspects of these trajectories, presenting the sequence of user inputs, robot actions, and amplified or error actions for each task. The user input is shown as a black line, representing commands such as moving left, right, or remaining neutral. These inputs are overlaid with the robot's actions, which amplifies the action space to additional movements like forward, backward, grasp, and release. The robot's amplified actions are also displayed, demonstrating instances where ARAS autonomously expanded upon user inputs to execute actions that users were limited to do such as precise alignment, moving forward, grasping, or releasing. This amplification is evident in the plots, where segments marked in green indicate robot actions taken without direct user commands, showcasing ARAS’s ability to infer and act upon user intent effectively. Additionally, error actions, marked in orange, highlight moments where ARAS deviated from the intended trajectory due to incorrect inference or external disturbances. This figure illustrates ARAS combining user control with autonomous decision-making, amplifying user inputs for efficient task completion while adapting to environmental changes. \ref{fig:trajfull}b shows ARAS's responsiveness to varying inputs, autonomously executing tasks when inputs are insufficient or imprecise, with minimal errors reflecting robust intent interpretation.

\begin{figure*}[!ht]
\centering
\includegraphics[width=7in]{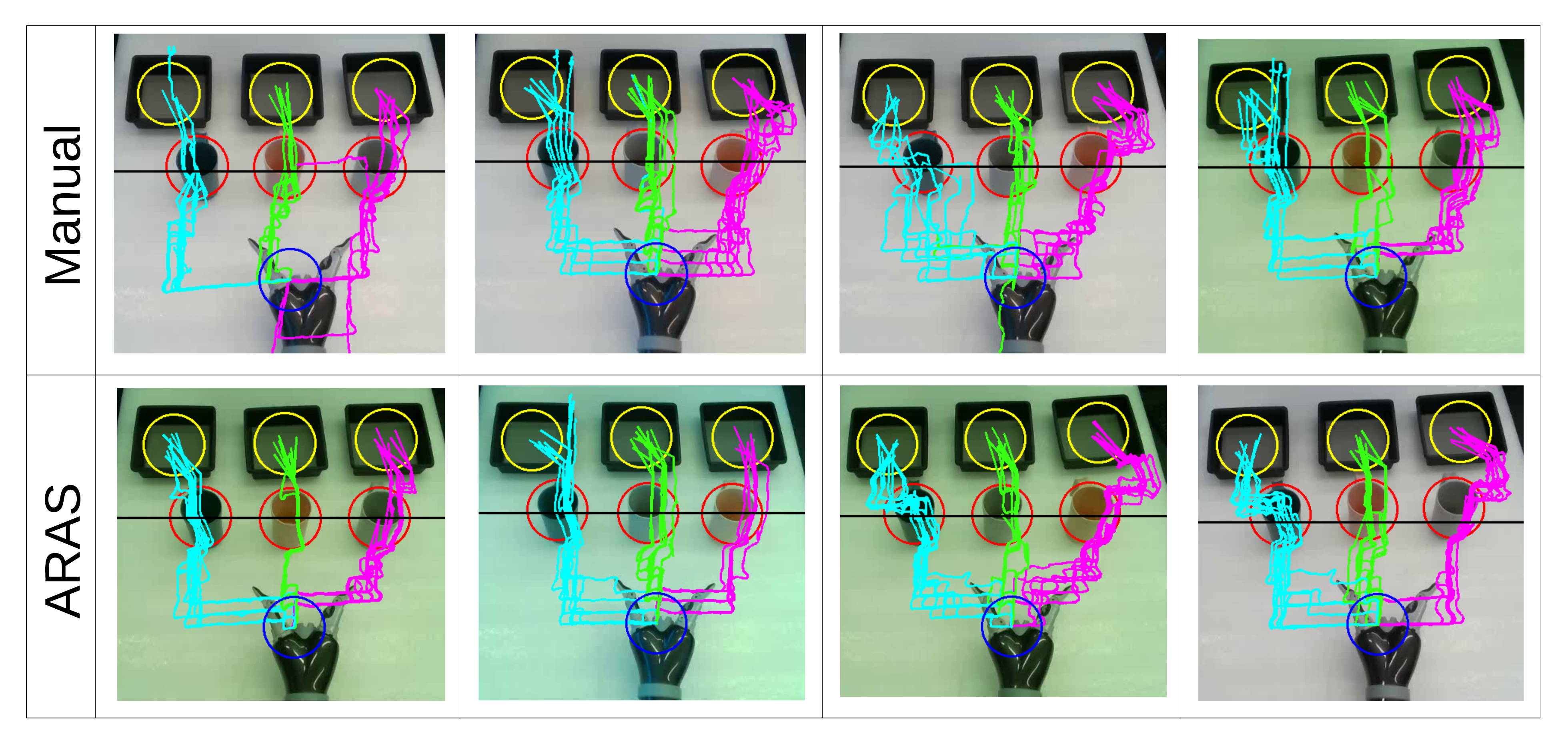}
\caption{Comparison of trajectories generated by manual mode (top row) and ARAS (bottom row) during pick-and-place tasks. The color-coded trajectories represent paths for three objects and bins.}
\label{fig:traj_all}
\end{figure*}

Figure \ref{fig:traj_all} compares the trajectories generated by the manual mode (top row) and ARAS (bottom row) during the pick-and-place tasks. In the manual mode, the trajectories exhibit noticeable variability, with inconsistent paths toward the goals, particularly for the rightmost bin (purple trajectories). The increased curvature and deviations in these trajectories are likely caused by the greater distance of the rightmost goal from the starting position and the participants' limited perception and control accuracy over longer distances. This variability indicates challenges in maintaining precise alignment with the objects and bins during task execution under manual control. The average trajectory spread for manual mode was calculated as 4.86 cm, reflecting the greater variability in user control.

In contrast, ARAS trajectories demonstrate significantly lower variance, with paths that are more streamlined and consistently aligned with the goals. The reduction in trajectory variability highlights ARAS's ability to amplify user inputs and autonomously optimize task execution, ensuring robustness and precision even for distant goals. The paths toward the rightmost bin under ARAS, while still reflecting some curvature due to the goal's distance, are notably more controlled compared to the manual mode, underscoring ARAS's capacity to mitigate challenges caused by user perception limitations. The trajectory spread can calculated as the standard deviation of the trajectory points relative to their mean position for each mode. The average trajectory spread for ARAS was 3.94 cm, significantly lower than the manual mode (p = 0.0019). This reduction illustrates the robustness of ARAS in maintaining consistent and precise trajectories.

While the overall trend for ARAS trajectories is similar to the manual mode, ARAS executes the tasks more robustly, reducing variability and optimizing alignment with the goals. This highlights its ability to effectively support users by amplifying their inputs and compensating for limitations in manual control. Overall, this figure illustrates how ARAS reduces trajectory variability and enhances goal-directed behavior, providing smoother, more reliable paths while accommodating the challenges associated with user input limitations and task complexity.

\begin{table*}[h!]
\centering
\caption{Comparison of Approaches in Assistive Robotics (IS: Input Space, D: Discrete, C: Continuous, I: Invasive, N: Non-invasive, T: Translational, R: Rotational)}
\label{tab:comparison}
\begin{tabular}{p{3cm}c c c c c c c}
\toprule
\textbf{Reference} & \textbf{Task} & \textbf{\shortstack{Mode\\switching}} & \textbf{\shortstack{Modality\\(IS, D/C, I/N)}} & \textbf{\shortstack{Control output\\DoF}} & \textbf{\shortstack{Dataset\\Req.}}& \textbf{\shortstack{Goal\\Awareness}} & \textbf{\shortstack{Dynamic\\Inference}} \\
\midrule
Muelling et al., 2017 \cite{muelling2017autonomy} & Pick and Place & No& Spikes (192, C, I) & 4(3T + grasp) & Yes & Estimated & Yes \\
Downey et al., 2016 \cite{downey2016blending} & Pick and Place & No& Spikes (240, C, I) & 4(3T + grasp) & Yes & Estimated & Yes \\
Natraj et al.,2024 \cite{natraj2023flexible} & Pick and Place & Yes & ECoG (128, C, I) & 6(3T + 2R + grasp) & Yes & Known& No \\
Hochberg et al., 2012 \cite{hochberg2012reach} & Self feeding & No& Spikes (96, C, I) & 5(3T + 1R + grasp) & Yes & Known & No \\
Xu et al., 2019 \cite{xu2019shared} & Reach and Grasp & Yes& EEG (32, C, N) & 2(1T + grasp) & Yes & Estimated & No \\
Beraldo et al., 2022 \cite{beraldo2022shared} & Navigation & No& EEG (16, C, N) & 1(T) & Yes & Estimated & No \\
Herlant et al., 2016 \cite{herlant2016assistive} & Object Manip. & Yes& Joystick (3, C, N) & 7(3T + 3R + grasp) & No & Known & No \\
Rudigkeit et al., 2014 \cite{rudigkeit2014towards} & Object Manip. & Yes& Head array (3, C, N) & 7(3T + 3R + grasp) & No & Known & No \\
Javdani et al., 2018 \cite{javdani2018shared} & Reach and Grasp & Yes& Joystick (2, C, N)& 7(3T + 3R + grasp) & No & Estimated & Yes \\
Yow et al., 2024 \cite{yow2023shared} & Reach and Grasp & Yes& Joystick (3, C, N) & 7(3T + 3R + grasp) & Yes & Estimated & Yes \\
\textbf{ARAS (ours)} & \textbf{Pick and Place} & \textbf{No}& \textbf{Keyboard (1, D, N)} & \textbf{4(3T + grasp)}& \textbf{No} & \textbf{Estimated} & \textbf{Yes} \\
\bottomrule
\end{tabular}
\end{table*}

\section{Discussion}
This study introduces ARAS, a multimodal shared autonomy framework designed to amplify low-dimensional user inputs for controlling high-dimensional robotic systems. The system was evaluated on a complete pick-and-place task, which includes both reaching and grasping as well as object placement. Unlike studies that focus solely on reach-and-grasp tasks, such as those by Xu et al. \cite{xu2019shared}, Javdani et al. \cite{javdani2018shared}, and Yow et al. \cite{yow2023shared}, ARAS addresses the full cycle of dexterous manipulation, which has more practical significance in real-world applications. A reach-and-grasp task alone lacks utility, as the act of grasping always requires a subsequent placement of the object to achieve a meaningful outcome. These prior studies, while effective in their respective scopes, often do not evaluate the placement phase, limiting their direct applicability to real-world scenarios where task completion requires both reaching and placing objects.

Table \ref{tab:comparison} provides a concise comparison of ARAS with other SOTA approaches in assistive robotics. The modalities used in these systems vary significantly, reflecting the trade-offs between invasiveness, input complexity, and task performance. Systems such as those by Natraj et al. \cite{natraj2023flexible}(our prior work), Muelling et al. \cite{muelling2017autonomy}, Downey et al. \cite{downey2016blending}, and Hochberg et al. \cite{hochberg2012reach} employ invasive neural signals like spikes or Electrocorticography (ECoG), which provide high-dimensional input but come with significant drawbacks, including clinical risks, ethical concerns, and limited scalability for broader applications. On the other hand, non-invasive systems, such as those by Javdani et al. \cite{javdani2018shared} and Rudigkeit et al. \cite{rudigkeit2014towards}, often rely on higher input dimensions coupled with mode-switching mechanisms to manage degrees of freedom. While effective, mode-switching introduces inefficiencies, with Herlant et al. \cite{herlant2016assistive} reporting that mode changes accounted for approximately 17.4\% of total task execution time. ARAS addresses this challenge by leveraging a multimodal latent space representation to amplify low-dimensional user inputs into high-dimensional robotic actions in real-time, enabling seamless control without the need for mode-switching. This approach not only improves efficiency but also simplifies the user interface while maintaining robust task performance.

Moreover, Many assistive robotics systems rely on extensive datasets for training, especially those using neural signals like spikes or EEG. These datasets often require time-consuming and invasive collection processes, as seen in approaches by Natraj et al. \cite{natraj2023flexible},  Muelling et al. \cite{muelling2017autonomy}, Downey et al. \cite{downey2016blending}, and Xu et al. \cite{xu2019shared}, which rely on high-dimensional neural inputs to model user intentions. This reliance poses significant challenges, including the need for invasive procedures, ethical concerns, and difficulty generalizing to diverse users or tasks. EEG-based systems, while non-invasive, still require substantial user training to ensure accurate signal decoding, further complicating practical deployment. ARAS addresses these limitations by synthesizing user inputs during training, generating noisy but realistic input signals to train the system in a simulation environment. This sim-to-real approach not only eliminates the need for extensive datasets but also ensures robustness by preparing the system for real-world variability in user inputs, enabling efficient and scalable deployment without relying on invasive or time-intensive data collection processes.

Goal awareness and adaptability to evolving user intentions are another critical aspect for effective shared autonomy. Many systems, such as Javdani et al. \cite{javdani2018shared}, operate under the assumption that user intentions remain fixed throughout the task, which limits their flexibility in dynamic or uncertain environments. While this predict-then-act approach works well for static goals, it struggles with tasks where user preferences change over time. On the other hand, approaches like Yow et al. \cite{yow2023shared} attempt to address evolving intentions through active information gathering during the task, such as querying users to refine goal predictions. However, this method introduces additional cognitive load for the user and increases task completion time, as the system must repeatedly pause for input. ARAS overcomes these limitations by employing a dynamic goal inference mechanism that continuously updates its understanding of the user’s intentions in real-time. This eliminates the need for explicit user input during the task, reducing cognitive load and ensuring efficient task completion, even in scenarios with evolving user goals, shown in Table \ref{tab:simulation_results}.

The results of ARAS were evaluated through simulation and user studies, highlighting its effectiveness across various metrics, including success rates, task completion times, and error actions. ARAS outperformed baseline methods such as Hindsight Optimization (HO) \cite{javdani2018shared} and DQN across four scenarios involving fixed and dynamic user intentions. For fixed user intentions, ARAS achieved a success rate of 99.76\%, a completion time of 31.48 seconds, and minimal error actions (4.24 seconds), compared to DQN's 88.44\% success rate and 35.62 seconds, and HO's 85.40\% success rate and 36.03 seconds. Under dynamic conditions, such as changes in user intentions during the pick-up, drop-off, or both phases, ARAS consistently outperformed the baselines, maintaining success rates above 96\%, while DQN and HO scored significantly lower, particularly in more complex scenarios. 

Although direct comparisons are challenging due to differences in experimental setups, ARAS’s performance metrics suggest a potential advantage over other methods reported in the literature. For example, Xu et al. \cite{xu2019shared} achieved an average success rate of 70\% for reach-and-grasp tasks with a non-invasive EEG-based system. Similarly, Muelling et al. \cite{muelling2017autonomy} achieved an average success rate of 91.7\% with an invasive method for grasping and lifting up objects. In comparison to noninvasive and invasive approaches, ARAS achieved a success rate of 92.88\%, slightly below the autonomous system (95.60\%) but higher than manual control (86.46\%), while maintaining a balance between efficiency and user involvement. In terms of completion times,  Yow et al. \cite{yow2023shared}, using a POMDP-based framework, reported a completion time of approximately 50 seconds for reach to grasp tasks requiring active information gathering, which increased cognitive load. In comparison, ARAS dynamically infers user intentions in real-time without the need for mode-switching or explicit user queries, achieving shorter task completion times, averaging between 30 to 40 seconds during user studies for a full pick-and-place task. Completion times for ARAS were faster than many comparable systems, and error actions were minimal, significantly lower than manual control, and comparable to the autonomous system. These results suggest that, while direct comparisons across studies are not entirely possible, ARAS potentially offers superior performance in terms of efficiency and adaptability across a broader range of tasks.

Furthermore, we investigate additional aspects and analyses in the Supplementary Materials, including the learning curves for both manual mode and ARAS. While manual mode exhibits a pronounced learning curve—requiring users to gradually improve their performance through practice—ARAS shows consistently high accuracy from the very beginning, highlighting its user-friendly design and minimal training requirements. Detailed statistical analyses and graphical representations in the supplementary figures and tables underscore this contrast, emphasizing that ARAS effectively eliminates the need for an extensive learning period. Additionally, a series of supplementary movies demonstrate the system's adaptability in both fixed and dynamically changing task scenarios, reinforcing the overall robustness and ease of use of our approach. 

\section{Conclusion}
This paper introduced ARAS, a multimodal shared autonomy framework that leverages a latent space representation and a deep reinforcement learning setup to amplify very limited user inputs while dynamically adapting to evolving user intentions in high-dimensional robotic tasks. By addressing challenges such as mode-switching, predefined assumptions, dataset dependency, and the requirement for high-dimensional inputs—which users with severe impairments may be unable to provide—ARAS offers a seamless and intuitive interface for effective user-robot collaboration. Evaluations on a complete pick-and-place task demonstrated its practical significance, achieving high success rates, reduced task completion times, and minimal error actions. ARAS outperformed baseline methods like Hindsight Optimization (HO) and DQN in both fixed and dynamic intention scenarios, highlighting its adaptability and efficiency.





\section{Acknowledgment}
The proposed research was supported by the NSF under award ID 2245558 and by the Rhode Island INBRE program from the National Institute of General Medical Sciences of the NIH under grant number P20GM103430. Additionally,  this project was partially supported by the URI Foundation Grant on Medical Research.



\ifCLASSOPTIONcaptionsoff
  \newpage
\fi

\printbibliography

\end{document}